\documentclass{article}
\usepackage[preprint]{wenxiaobai}
\usepackage[utf8]{inputenc} % allow utf-8 input
\usepackage[T1]{fontenc}    % use 8-bit T1 fonts
\usepackage{hyperref}       % hyperlinks
\usepackage{url}            % simple URL typesetting
\usepackage{booktabs}       % professional-quality tables
\usepackage{amsfonts}       % blackboard math symbols
\usepackage{nicefrac}       % compact symbols for 1/2, etc.
\usepackage{microtype}      
\usepackage{etoolbox}
\usepackage{caption}
\usepackage{adjustbox}

\usepackage{amsmath} % 确保加载 amsmath
\usepackage{amssymb} % 通常也需要，提供 \mathcal 等
\usepackage{multirow}
\usepackage{siunitx}
\usepackage{nicematrix}
\usepackage{enumitem}
\usepackage{tabularx}
\usepackage{hyphenat}
\tcbuselibrary{breakable} % To allow boxes to break across pages
\usepackage{fancyvrb}    % For the Verbatim environment (capital 'V')
\tcbuselibrary{breakable,skins}   % “enhanced” 位于 skins 库中
\tcbuselibrary{listings}          % 可选：让 tcolorbox 里支持 verbatim/listing
\usepackage{enumitem}  % For customizing lists
\usepackage{hyperref}  % For creating hyperlinks
\usepackage{fvextra}   % For line breaking in verbatim

\DefineVerbatimEnvironment{Verbatim}{Verbatim}{breaklines=true, breakanywhere=true}

\definecolor{ysdarkpurple}{HTML}{4E2399}
\definecolor{ysshallowpurple}{HTML}{E6DBFF}
\definecolor{ysdarkred}{HTML}{8c2824}
\definecolor{ysshallowred}{HTML}{F8D7D7}
\definecolor{ysdarkblue}{HTML}{005E99}
\definecolor{ysshallowblue}{HTML}{CCEBFF}
\definecolor{ysdarkgrey}{HTML}{333333}
\definecolor{ysshallowgrey}{HTML}{E5E5E5}
\usepackage{tabularx, array}
\newcolumntype{Y}{>{\centering\arraybackslash}X} % 居中可伸缩列

\definecolor{ColorGrok}{HTML}{FFFDE7}      % 保持不变
\definecolor{ColorPplx}{HTML}{EFFDFE}      % 保持不变
\definecolor{ColorOpenAI}{HTML}{F2F2F2}    % 将 EDEDED 进一步调浅为 F2F2F2 (非常浅的灰色)
\definecolor{ColorGemini}{HTML}{E6F4FE}    % 保持不变
\definecolor{ColorClaude}{HTML}{FFF3EB}    % 将 FFF0E5 进一步调浅为 FFF3EB (更极浅的桃色/米橙色)
\definecolor{SectionHeaderColor}{HTML}{FFFFFF} % 保持不变 (白色)

\colorlet{DarkerColorClaude}{ColorClaude!95!black}
\colorlet{DarkerColorPplx}{ColorPplx!95!black}
\colorlet{DarkerColorGemini}{ColorGemini!95!black}
\colorlet{DarkerColorOpenAI}{ColorOpenAI!95!black}
\colorlet{DarkerColorGrok}{ColorGrok!95!black}
\graphicspath{{figures/}}

\title{DeepResearch Bench: A Comprehensive Benchmark for \\
Deep Research Agents}

\author{
{\bf Mingxuan Du\textsuperscript{\rm 1}\thanks{Work done during the internship at Metastone.}\, , Benfeng Xu\textsuperscript{\rm 1,2}, Chiwei Zhu\textsuperscript{\rm 1}, Xiaorui Wang\textsuperscript{\rm 2}, Zhendong Mao\textsuperscript{\rm 1}\thanks{Corresponding author: Zhendong Mao.}} \\
\textsuperscript{1}University of Science and Technology of China,
\textsuperscript{2}MetastoneTechnology, Beijing, China \\ 
\texttt{\{dumingxuan, benfeng\}@mail.ustc.edu.cn}
}

\begin{document}

\maketitle

\begin{abstract}
    Deep Research Agents are a prominent category of LLM-based agents. By autonomously orchestrating multistep web exploration, targeted retrieval, and higher-order synthesis, they transform vast amounts of online information into analyst-grade, citation-rich reports---compressing hours of manual desk research into minutes. However, a comprehensive benchmark for systematically evaluating the capabilities of these agents remains absent. To bridge this gap, we present \textbf{DeepResearch Bench}, a benchmark consisting of 100 PhD-level research tasks, each meticulously crafted by domain experts across 22 distinct fields. Evaluating DRAs is inherently complex and labor-intensive. We therefore propose two novel methodologies that achieve strong alignment with human judgment. The first is a reference-based method with adaptive criteria to assess the quality of generated research reports. The other framework is introduced to evaluate DRA's information retrieval and collection capabilities by assessing its effective citation count and overall citation accuracy. We have open-sourced DeepResearch Bench and key components of these frameworks at \href{https://github.com/Ayanami0730/deep_research_bench}{https://github.com/Ayanami0730/deep\_research\_bench} to accelerate the development of practical LLM-based agents.
\end{abstract}

\begin{figure}[h!] 
  \centering
  \captionsetup{skip=2pt}
  \includegraphics[width=0.99\textwidth]{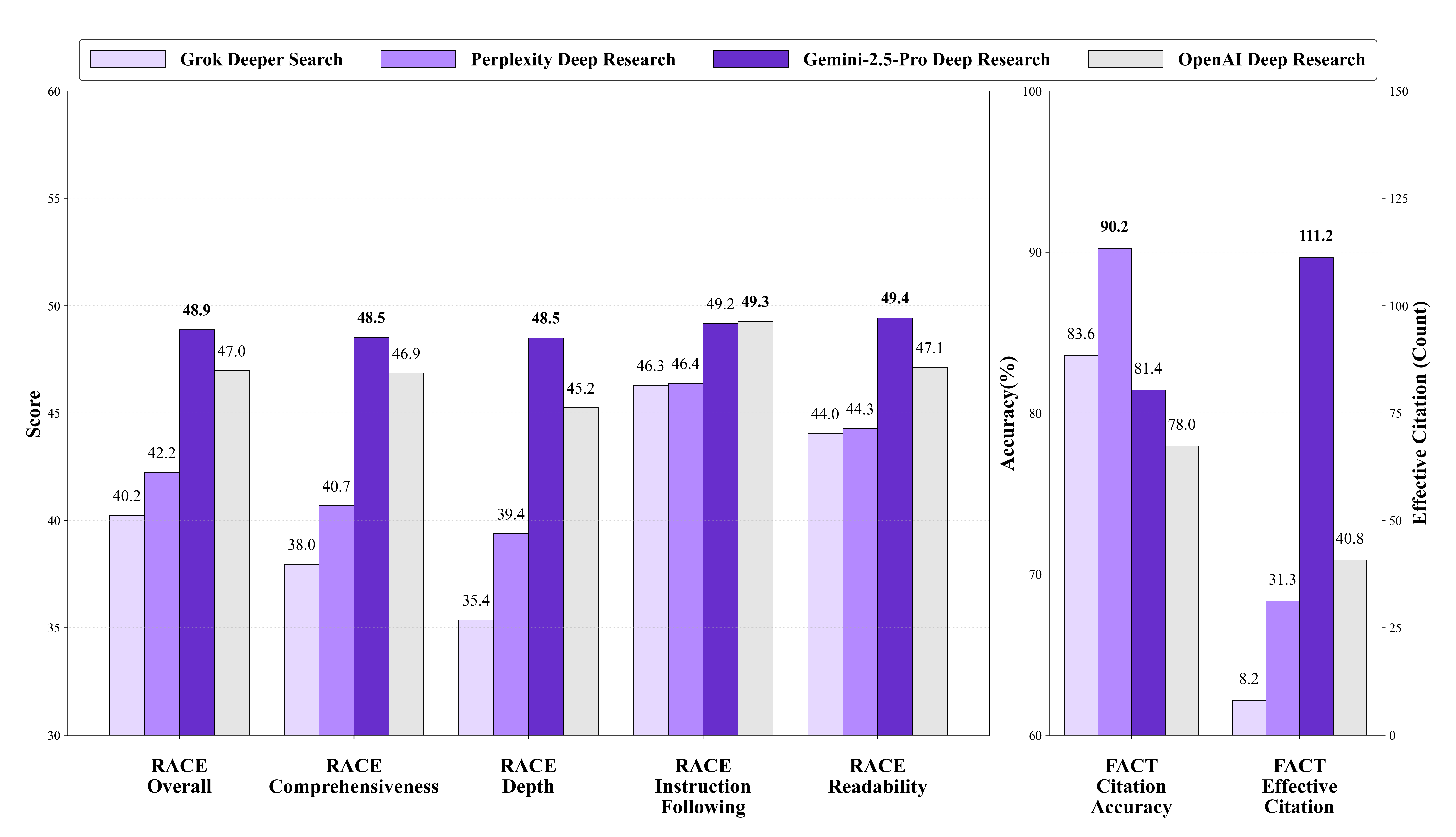}
  \caption{Overview of agent performance on DeepResearch Bench. \textbf{Left}: Generated report quality scores across evaluation dimensions. \textbf{Right}: Agent citation accuracy and average number of effective citations.}
  \label{fig:agent_performance_overview} % A new label for this figure
\end{figure}

\newgeometry{top=1.2in, textheight=8.7in, textwidth=6.5in, headheight=12pt, headsep=25pt, footskip=30pt}

\section{Introduction}
We now enter a new phase of AI \cite{yao_second_nodate}, a period marked by comprehensive advances in the capabilities of the large language model (LLM) \cite{deepseek-ai_deepseek-r1_2025, openai_openai_2024}. These advancements enable the construction of LLM-based Agent systems designed to tackle increasingly complex tasks \cite{masterman_landscape_2024, hong_metagpt_2024, yang_swe-agent_2024}. In this evolving landscape, defining tasks that genuinely reflect real-world demands and designing robust evaluation methodologies to measure the progress of these Agent systems are becoming critically important. Deep research represents one such well-defined task domain, with Deep Research Agents (DRAs) \cite{li_webthinker_2025, zheng_deepresearcher_2025, schmidgall_agent_2025} emerging as the most widely utilized LLM-based agents today. Users leverage these agents to enhance their productivity significantly.

However, comprehensively evaluating the capabilities of these DRAs presents substantial challenges. Because their internal reasoning and information retrieval processes are not transparent, the final generated report becomes the primary interface through which their overall performance can be assessed. Moreover, evaluating the quality of these extensive research reports remains an open-ended problem, as establishing a definitive 'ground truth' or standard answer for complex research queries is almost impossible.

These demanding evaluation requirements for DRAs pose a significant challenge for existing evaluation frameworks, which often fall short of offering a dedicated assessment of the multifaceted capabilities of such agents \cite{liu_agentbench_2023}. Current benchmarks typically focus on assessing isolated capabilities—such as web browsing and information retrieval \cite{wei_browsecomp_2025, zhou_browsecomp-zh_2025, zhou_webarena_2024}, or generative abilities disconnected from real-time information acquisition \cite{que_hellobench_2024, bai_longwriter_2024, wu_writingbench_2025}.

To fill this gap in evaluating end‑to‑end research agents, we introduce \textbf{DeepResearch Bench}, a 100‑task benchmark spanning 22 domains, with each task meticulously crafted and iteratively refined by domain experts. To ensure this benchmark authentically reflects real research needs, we carefully determine the specific number of tasks from each topic domain. This is guided by a statistical analysis of over 96,000 real-world user queries, adhering to the pipeline illustrated in Figure~\ref{fig:framework}(a).

Building on this dataset, we introduce two novel, highly human-aligned evaluation methodologies to comprehensively assess both the quality of final reports generated by DRAs and their ability to gather web information effectively. The first one is a \underline{R}eference-based and \underline{A}daptive \underline{C}riteria-driven \underline{E}valuation framework with Dynamic Weighting (denoted as \textbf{RACE} for ease of subsequent reference), which targets the assessment of report generation quality. And the other one is a framework for \underline{F}actual \underline{A}bundance and \underline{C}itation \underline{T}rustworthiness (denoted as \textbf{FACT}), which focuses on evaluating information retrieval and citation accuracy. Overview results are shown in Figure~\ref {fig:agent_performance_overview}. \textbf{Furthermore, we believe these methodologies are not confined to deep research scenarios}. RACE can serve as a general framework for evaluating text generation capabilities, while FACT offers a robust approach for assessing the coverage and reliability of information retrieved by LLMs.

Our primary contributions are as follows:

\begin{itemize}[leftmargin=20pt]
\item We present \textbf{DeepResearch Bench}, the first specialized benchmark for evaluating Deep Research Agents, constructed through a systematic process that reflects authentic user needs.
\item We further propose \textbf{RACE} and \textbf{FACT}, two novel evaluation frameworks that respectively assess the report generation quality and the information retrieval abilities of Deep Research Agents.
\item We conduct comprehensive human studies to validate the reliability of our frameworks and publicly release the benchmark and evaluation protocols to foster future research.
\end{itemize}

\section{DeepResearch Bench: Topic Distribution Analysis \& Data Collection}
\label{sec:DeepResearch Bench}

\subsection{Topic Distribution Analysis}
\label{subsec: distribution analysis}

Deep Research Agents (DRAs) are intended to serve actual human research needs. Therefore, to effectively test their capabilities, the design of DeepResearch Bench should be grounded in the real-world distribution of human research task demands. To obtain this distribution, we collect an in-house dataset of 96,147 raw user queries from interactions with web search-enabled LLM Chatbot. To ensure user privacy, all raw query logs underwent rigorous anonymization, removing personally identifiable information such as user IDs, IP addresses, and session metadata. 

We then define the concept of Deep Research tasks as problems requiring agents to conduct multiple rounds of web searches, gather information, perform analysis, and produce high-quality reports. Guided by this definition, we employ DeepSeek-V3-0324 \cite{deepseek-ai_deepseek-aideepseek-v3-0324_2025} to conduct filtering, identifying queries that aligned with our deep research requirements. This process ultimately yields a dataset of 44,019 queries conforming to our deep research task definition.

\begin{figure}[t]
\centering
\includegraphics[width=0.99\textwidth]{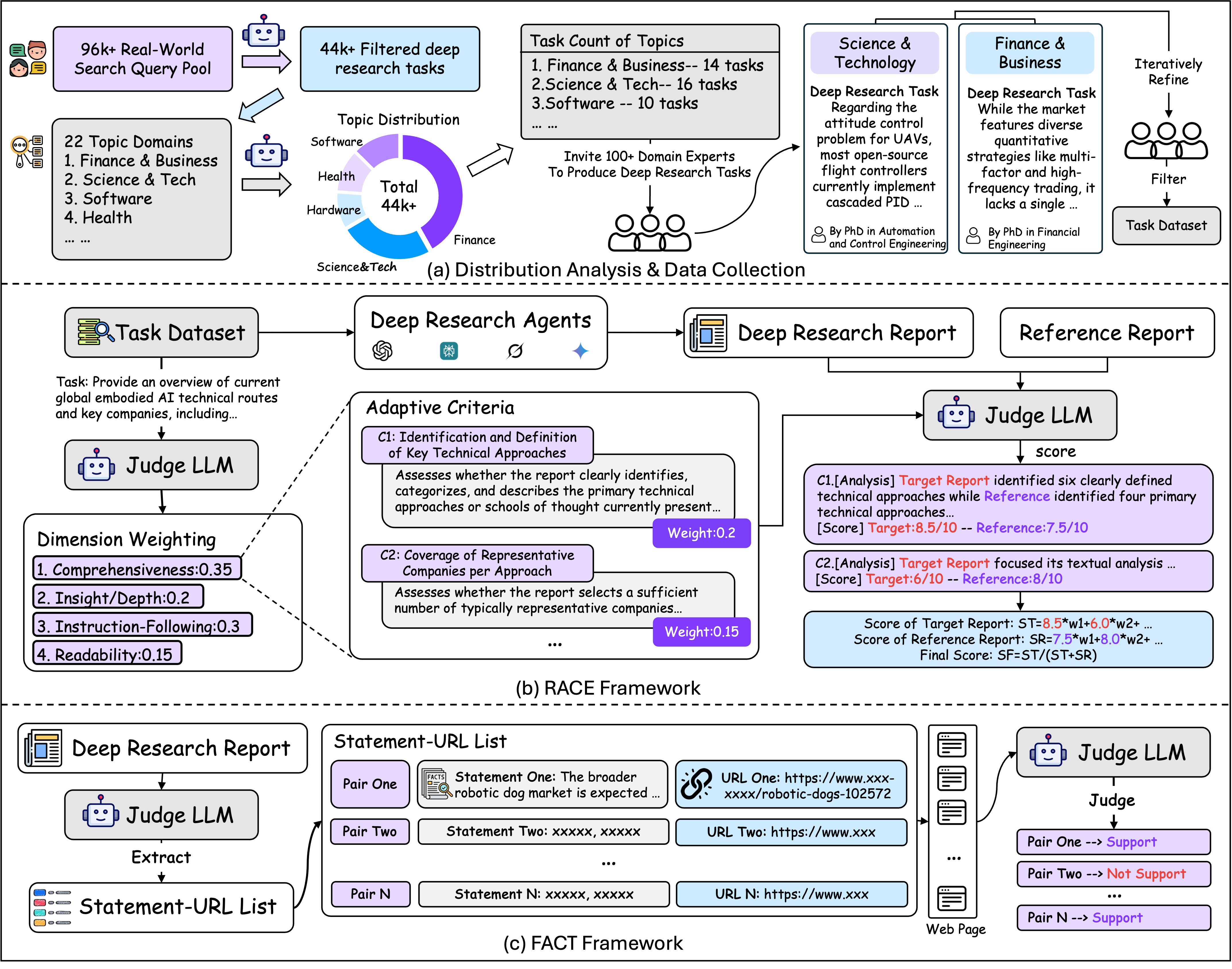}
\caption{Overview of DeepResearch Bench. (a)Distribution analysis and dataset construction pipeline. 
(b)Overview of RACE(a \underline{R}eference-based \underline{A}daptive \underline{C}riteria-driven \underline{E}valuation framework with Dynamic Weighting)
(c)Overview of FACT (a Framework for \underline{F}actual \underline{A}bundance and \underline{C}itation \underline{T}rustworthiness)}
\label{fig:framework}
\end{figure}

\begin{figure}[ht]
\centering
\includegraphics[width=0.99\textwidth]{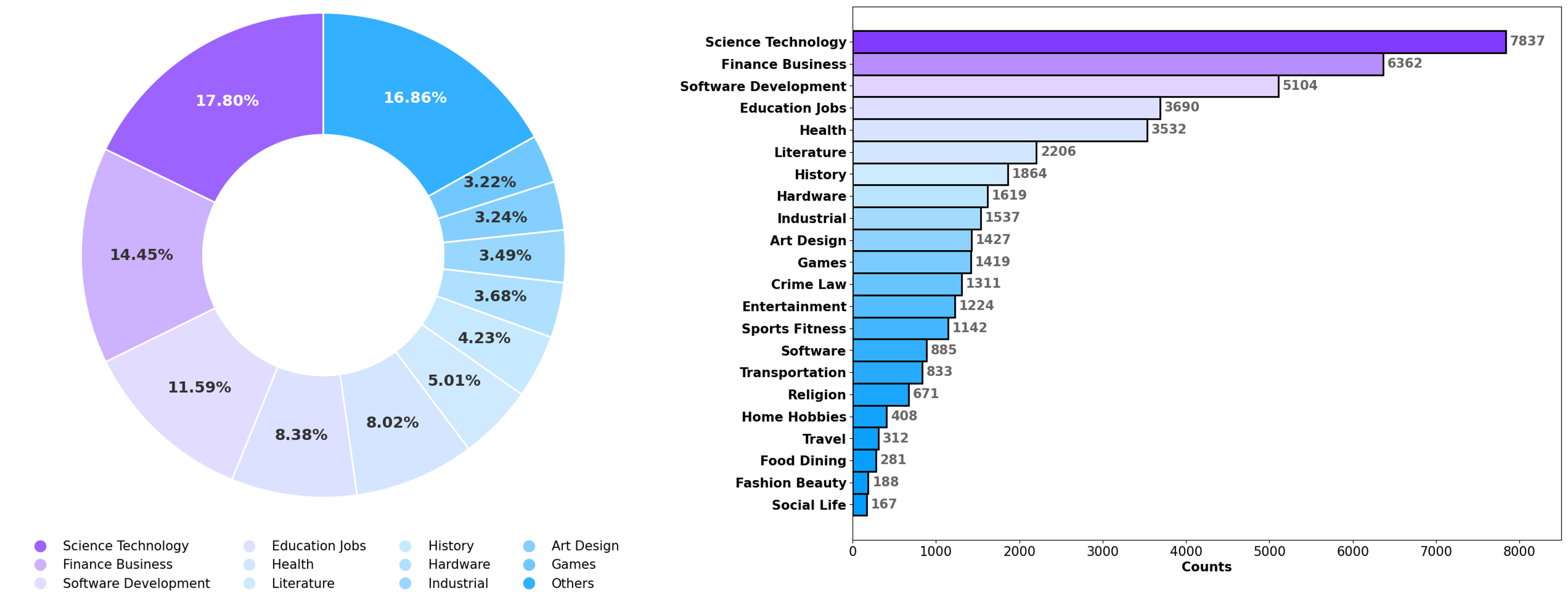}
\caption{Topic distribution of the 44,019 filtered deep-research tasks.  
\textbf{Left:} Donut chart showing the proportional share of each topic domain; the \emph{Others} slice aggregates the least-represented categories.  
\textbf{Right:} Bar chart listing the absolute task counts for all 22 domains.}
\label{fig:topic_distribution}
\end{figure}

To categorize the deep research queries, we adopt the topic taxonomy proposed by WebOrganizer \cite{wettig_organize_2025}, selecting 22 distinct topic domains for this classification. We then employ DeepSeek-V3-0324 to classify these 44,019 queries into the 22 selected topic domains. By statistically aggregating the LLM's classification results, we obtain the distribution of these queries across the various topics. This distribution, visualized in Figure~\ref{fig:topic_distribution}, indicates the real-world user demand for deep research within these domains.

\subsection{Benchmark Task Collection}
\label{subsec: task collection}

Guided by the real-world user demand distribution, we first determine the target number of tasks for each topic domain within the DeepResearch Bench dataset. Subsequently, considering the significant resources required to run deep research agents and evaluate their results, we proportionally compress this determined distribution to form a final dataset of 100 tasks, 50 in Chinese and 50 in English. This compression process meticulously ensures that the benchmark maintains the same topical balance as the observed real-world distribution.

\begin{figure}[ht]
\centering
\includegraphics[width=0.99\textwidth]{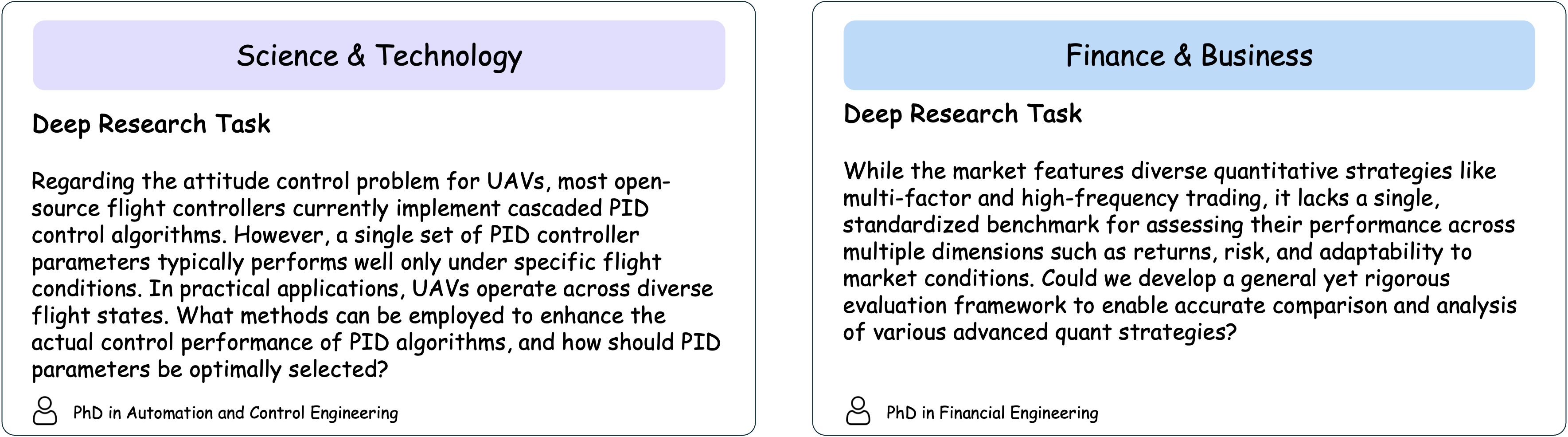}
\caption{Two example tasks from DeepResearch Bench.}
\label{fig:showcase}
\end{figure}

Once the target task count for the topic domain is determined, our focus shifts to constructing research tasks that are highly challenging and firmly grounded in authentic research demands. This process specifically aims to test the upper limits of the Deep Research Agents' capabilities. We invite Ph.D. or senior practitioners with more than five years of relevant domain experience to propose candidate tasks, as shown in Figure~\ref{fig:showcase}. All submissions undergo a manual screening by our team to verify their quality, clarity, complexity, and alignment with our definition of deep research. This rigorous vetting process yields the 100 high-quality benchmark tasks that comprise the DeepResearch Bench.

\section{Evaluation Methodology}
\label{sec:evaluation}

Our evaluation methodology focuses on two critical aspects: their capabilities in information retrieval and collection, as well as the quality of the final reports they generate. To assess these respective dimensions, we developed two complementary frameworks, RACE and FACT.

\subsection{RACE: A Framework for Report Quality Evaluation}
\label{subsec:race_framework}

Evaluating long-form research reports presents significant challenges. Existing approaches using fixed checklists \cite{que_hellobench_2024} or static rubrics \cite{shao_assisting_2024,bai_longwriter_2024} struggle to adapt to diverse tasks, specialized domains, and nuanced quality aspects of deep research tasks. To address this, we introduce our \underline{R}eference-based \underline{A}daptive \underline{C}riteria-driven \underline{E}valuation framework with Dynamic Weighting (\textbf{RACE}), leveraging the LLM-as-a-Judge method \cite{zheng_judging_2023}.  RACE offers a more adaptive and robust evaluation by first dynamically generating task-specific weights and criteria. It then employs a reference-based scoring approach, comparing the target report with a high-quality reference. Finally, a relative score is computed to assess the quality of the target report.

\paragraph{Dynamic Weight \& Adaptive Criteria Generation.}
Directly prompting LLMs to generate task-specific criteria from scratch can lead to results that deviate significantly from the intended assessment goals. Therefore, we first establish four top-level orthogonal dimensions based on domain expertise: \textbf{Comprehensiveness (\textsc{Comp})}, \textbf{Insight/Depth (\textsc{Depth})}, \textbf{Instruction-Following (\textsc{Inst})}, and \textbf{Readability (\textsc{Read})}. Detailed definitions are provided in the Appendix~\ref{app:dimension_definitions}. As illustrated in Figure~\ref{fig:framework}(b), given task $t$, the Judge LLM derives task-specific dimension weights by averaging weights $w_d^{(j)}$ obtained over $T$ trials for each dimension $d \in \{\textsc{Comp}, \textsc{Depth}, \textsc{Inst}, \textsc{Read}\}$:

\begin{equation}
W_d = \frac{1}{T} \sum_{j=1}^{T} w_{d}^{(j)}
\end{equation}

These final weights, $W_d$, ensure that evaluation aligns with the task's intent. Subsequently, for each dimension $d$, the Judge LLM generates a set of $K_d$ tailored criteria $\{c_{d,k}\}$ and their corresponding weights $\{w_{d,k}\}$ (where $\sum_{k=1}^{K_d} w_{d,k} = 1$), which are clear and actionable for evaluating the report within that dimension.

\paragraph{Reference-Based Scoring.}
Preliminary experiments suggest that scoring reports in isolation often yields insufficiently discriminative results; models tend to assign uniformly high scores, thereby masking genuine quality variations. To mitigate this, RACE adopts a reference-based scoring strategy. For each task $t$, a high-quality report $R_{\mathrm{ref}}$ is selected as a reference. All generated criteria $\{c_{d,k}\}$ across all dimensions are aggregated into a comprehensive list $\mathcal{C}_t$. The Judge LLM then analyzes the target report $R_{\mathrm{tgt}}$ and the reference report $R_{\mathrm{ref}}$ against each criterion $c \in \mathcal{C}_t$. This yields lists of scores for both reports for each criterion, which are then used for final score calculation:

\begin{equation}
(\{s_{\mathrm{tgt}, c}\}_{c \in \mathcal{C}_t}, \{s_{\mathrm{ref}, c}\}_{c \in \mathcal{C}_t}) = \mathrm{JudgeLLM}(t, R_{\mathrm{tgt}}, R_{\mathrm{ref}}, \mathcal{C}_t)
\end{equation}

\paragraph{Overall Score Calculation.}
Finally, we compute the overall quality score of the target report. First, dimension-level scores $S_d(R)$ are calculated by weighting criterion-level scores $s_{R, c_{d,k}}$ with criterion weights $w_{d,k}$. Second, these $S_d(R)$ scores are weighted by dimension weights $W_d$ to produce intermediate overall scores $S_{\mathrm{int}}(R)$. Finally, the target report's score $S_{\mathrm{final}}(R_{\mathrm{tgt}})$ is determined relative to the reference report's score:
\begin{equation}
S_{\mathrm{final}}(R_{\mathrm{tgt}}) = \frac{S_{\mathrm{int}}(R_{\mathrm{tgt}})}{S_{\mathrm{int}}(R_{\mathrm{tgt}}) + S_{\mathrm{int}}(R_{\mathrm{ref}})}
\end{equation}

\subsection{FACT: A Framework for Web Retrieval Evaluation}
\label{subsec:fact_framework}
To assess the factual grounding of report content and the agent's effectiveness in retrieving and utilizing web-based information, we introduce the framework for \underline{F}actual \underline{A}bundance and \underline{C}itation \underline{T}rustworthiness(\textbf{FACT}). This framework evaluates DRAs through the following automated steps:

\paragraph{Statement-URL Pair Extraction and Deduplication.}
We employ a Judge LLM to extract discrete statements from reports generated by DRAs with their corresponding cited source URLs. Then, the Judge LLM examines the pairs to identify cases where multiple statements associated with the same URL describe exactly the same fact. In such cases, only one representative Statement-URL pair is retained, ensuring each unique factual claim is represented only once.

\paragraph{Support Judgment.}
Each unique Statement-URL pair undergoes a support evaluation. We retrieve the textual content of the webpage using the Jina Reader API. Then, the Judge LLM assesses whether this content provides sufficient evidence to support the statement. This results in a binary judgment ('support' or 'not support') for each pair, determining whether the citation accurately grounds the claim.

\paragraph{Calculation of Citation Metrics.}
Based on these support judgments, we calculate two primary metrics. \textbf{Citation Accuracy (C. Acc.)} measures the precision of an agent's citations, reflecting its ability to ground statements with appropriate sources correctly. And \textbf{Average Effective Citations per Task (E. Cit.)} quantifies the average amount of valuable, verifiably supported information an agent retrieves and presents per task. For detailed calculation methodologies, please refer to Appendix~\ref{app:citation_metrics_calculation}.

\section{Experiments}
\label{sec:experiments}

\subsection{Experimental Setup}
\label{subsec:setup}

\paragraph{Implementation Details}
When employing the RACE framework, an initial pre-processing step involves cleaning citation formatting from the generated reports, as overly lengthy or complex citation styles can adversely affect the Judge LLM's scoring process. For RACE evaluation tasks, we utilize \texttt{Gemini-2.5-pro} as the Judge LLM. As for the FACT framework, \texttt{Gemini-2.5-flash} is employed for both Statement-URL pair extraction and support judgment, which is sufficient in capabilities while more economical for the token-consuming citation verification task. The reference reports ($R_{\mathrm{ref}}$) used in RACE's scoring methodology were selected from deep research articles generated by the Gemini-2.5-pro-based Deep Research, as available in April 2025\footnote{We cannot confirm if there were any Deep Research model iterations during the reference collection period.}.

\paragraph{Evaluated Models}
In our work, we primarily evaluate four early-released Deep Research Agents: Gemini-2.5-pro-based Deep Research, OpenAI Deep Research, Grok Deeper Search, and Perplexity Deep Research. Due to the lack of transparency regarding the iteration cycles of these commercial products, we specify the data collection timeframe in Appendix~\ref{app:data_collection_timeframes}.. 
Additionally, we evaluate several other models that are not specialized Deep Research Agents. These include leading LLMs that provide built-in web search tools. Furthermore, for all such tested LLMs with search tools, we standardize evaluation conditions by setting the search\_context\_size to high. More details on these configurations can be found in Appendix~\ref{subsec:llm_search_config}.

\begin{table}[htbp]
  \centering
  \small                       % 如需与正文完全同字号，可删掉
  \setlength{\tabcolsep}{3.5pt}% 左右内边距
  \caption{Overall evaluation results of DeepResearch Bench.
           \textbf{Bold} denotes the highest score in each column for Deep Research Agents
           (and for LLM with Search Tools within their respective section).
           \underline{Underlined} denotes the second highest.}
  \label{tab:main_results}

  % ----------- 关键：tabularx + 7 个 Y 列 ---------------
  \begin{tabularx}{\textwidth}{l *{7}{Y}}
    \toprule
    \multirow{2}{*}{\textbf{Model}}
      & \multicolumn{5}{c}{\textbf{RACE}}
      & \multicolumn{2}{c}{\textbf{FACT}} \\
    \cmidrule(lr){2-6}\cmidrule(lr){7-8}
      & \textbf{Overall} & \textbf{Comp.} & \textbf{Depth}
      & \textbf{Inst.} & \textbf{Read.}
      & \textbf{C. Acc.} & \textbf{E. Cit.} \\
    \midrule
    \multicolumn{8}{c}{\emph{LLM with Search Tools}} \\
    \midrule
    \rowcolor{ColorClaude}
      Claude-3.7-Sonnet w/Search & \cellcolor{DarkerColorClaude}{\textbf{40.67}}
      & \textbf{38.99} & \textbf{37.66} & \textbf{45.77} & 41.46
      & \underline{93.68} & \underline{32.48} \\
    \rowcolor{ColorClaude}
      Claude-3.5-Sonnet w/Search & \cellcolor{DarkerColorClaude}{28.48}
      & 24.82 & 22.82 & 35.12 & 35.08 & \textbf{94.04} & 9.78 \\
    \rowcolor{ColorPplx}
      Perplexity-Sonar-Reasoning-Pro(high) & \cellcolor{DarkerColorPplx}{\underline{40.22}}
      & \underline{37.38} & 36.11 & \underline{45.66} & \textbf{44.74}
      & 39.36 & 8.35 \\
    \rowcolor{ColorPplx}
      Perplexity-Sonar-Reasoning(high) & \cellcolor{DarkerColorPplx}{40.18}
      & 37.14 & \underline{36.73} & 45.15 & \underline{44.35}
      & 48.67 & 11.34 \\
    \rowcolor{ColorPplx}
      Perplexity-Sonar-Pro(high) & \cellcolor{DarkerColorPplx}{38.93}
      & 36.38 & 34.26 & 44.70 & 43.35 & 78.66 & 14.74 \\
    \rowcolor{ColorPplx}
      Perplexity-Sonar(high) & \cellcolor{DarkerColorPplx}{34.54}
      & 30.95 & 27.51 & 42.33 & 41.60 & 74.42 & 8.67 \\
    \rowcolor{ColorGemini}
      Gemini-2.5-Pro-Grounding & \cellcolor{DarkerColorGemini}{35.12}
      & 34.06 & 29.79 & 41.67 & 37.16 & 81.81 & \textbf{32.88} \\
    \rowcolor{ColorGemini}
      Gemini-2.5-Flash-Grounding & \cellcolor{DarkerColorGemini}{32.39}
      & 31.63 & 26.73 & 38.82 & 34.48 & 81.92 & 31.08 \\
    \rowcolor{ColorOpenAI}
      GPT-4o-Search-Preview (high) & \cellcolor{DarkerColorOpenAI}{35.10}
      & 31.99 & 27.57 & 43.17 & 41.23 & 88.41 & 4.79 \\
    \rowcolor{ColorOpenAI}
      GPT-4o-Mini-Search-Preview(high) & \cellcolor{DarkerColorOpenAI}{31.55}
      & 27.38 & 22.64 & 40.67 & 39.91 & 84.98 & 4.95 \\
    \rowcolor{ColorOpenAI}
      GPT-4.1 w/Search(high) & \cellcolor{DarkerColorOpenAI}{33.46}
      & 29.42 & 25.38 & 42.33 & 40.77 & 87.83 & 4.42 \\
    \rowcolor{ColorOpenAI}
      GPT-4.1-mini w/Search(high) & \cellcolor{DarkerColorOpenAI}{30.26}
      & 26.05 & 20.75 & 39.65 & 39.33 & 84.58 & 4.35 \\
    \midrule
    \multicolumn{8}{c}{\emph{Deep Research Agent}} \\
    \midrule
    \rowcolor{ColorGrok}
      Grok Deeper Search & \cellcolor{DarkerColorGrok}{40.24}
      & 37.97 & 35.37 & 46.30 & 44.05 & \underline{83.59} & 8.15 \\
    \rowcolor{ColorPplx}
      Perplexity Deep Research & \cellcolor{DarkerColorPplx}{42.25}
      & 40.69 & 39.39 & 46.40 & 44.28 & \textbf{90.24} & 31.26 \\
    \rowcolor{ColorGemini}
      Gemini-2.5-Pro Deep Research & \cellcolor{DarkerColorGemini}{\textbf{48.88}}
      & \textbf{48.53} & \textbf{48.50} & \underline{49.18} & \textbf{49.44}
      & 81.44 & \textbf{111.21} \\
    \rowcolor{ColorOpenAI}
      OpenAI Deep Research & \cellcolor{DarkerColorOpenAI}{\underline{46.98}}
      & \underline{46.87} & \underline{45.25} & \textbf{49.27} & \underline{47.14}
      & 77.96 & \underline{40.79} \\
    \bottomrule
  \end{tabularx}
\end{table}

\subsection{Main Results}
\label{subsec:main_results}

\subsubsection{Evaluation on RACE Framework}
As shown in Table~\ref{tab:main_results}, within the DRA category, Gemini-2.5-Pro Deep Research demonstrates leading overall performance on the RACE framework. OpenAI Deep Research also exhibits strong capabilities, surpassing Gemini-2.5-Pro Deep Research in the Instruction-Following dimension. This indicates that the different dimensions are somewhat decoupled and that a sufficiently powerful Judge LLM can capture performance variations across these various dimensions. The scores for these top agents are relatively close, which is characteristic of the reference-based relative scoring employed by RACE. However, this should not be concerning, as subsequent experiments~\ref{subsec:human_consistency} reveal a strong linear correlation between these RACE scores and human judgments, suggesting the framework effectively captures meaningful performance differences between models. In fact, these scores are highly linearly correlated with human evaluations, mapped to a different reference frame (similar to how scores of 45 and 50 are compared to human scores of 90 and 100). Therefore, we should focus on the rankings and proportional differences between the scores rather than the absolute score values. It can be observed that OpenAI Deep Research and Gemini-2.5-Pro Deep Research show comparable performance, both significantly outperforming the other two Deep Research Agents. And Perplexity Deep Research slightly outperforms Grok Deeper Search.

\begin{figure}[t]
\centering
\includegraphics[width=0.99\textwidth]{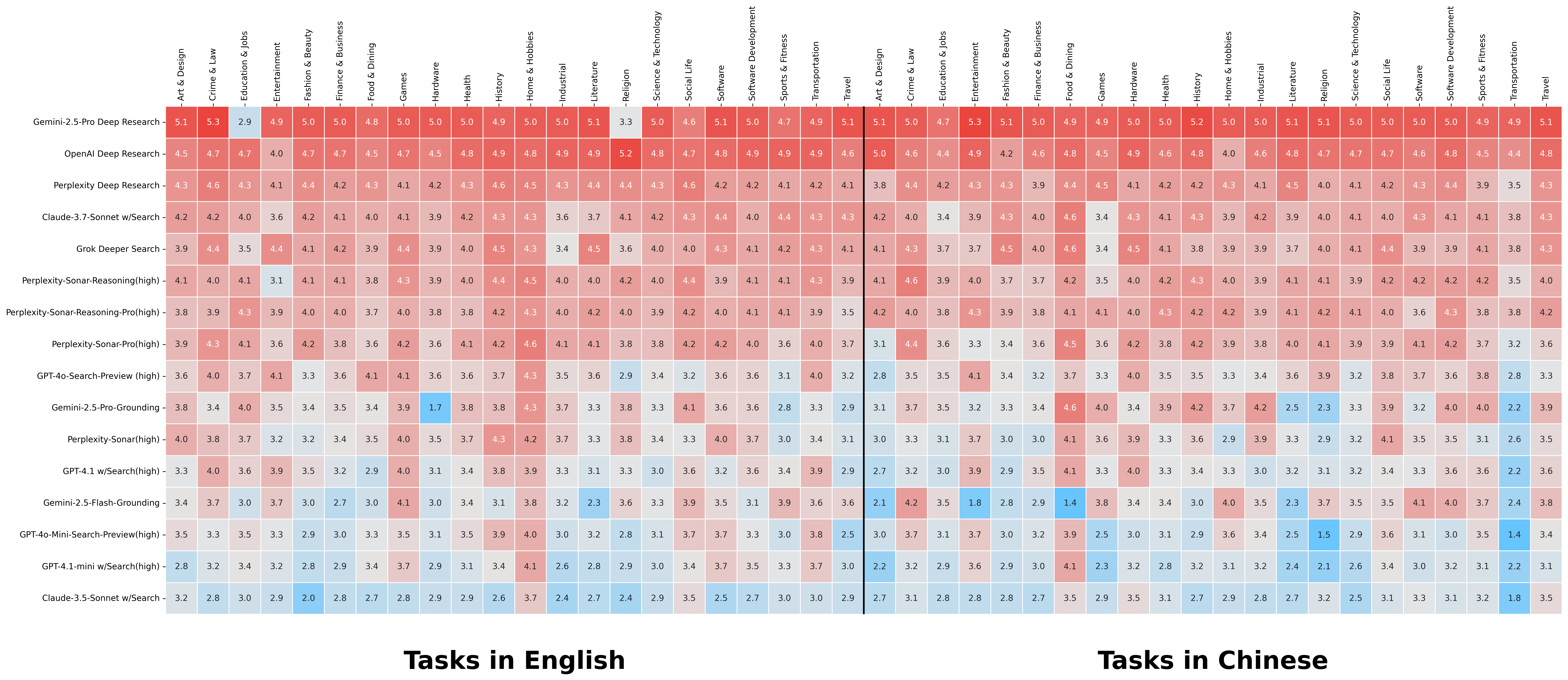}
\caption{Scores of different models across various topics and languages evaluated by RACE. Red indicates a higher score while blue refers to a lower score.}
\label{fig:topic_score_heatmap}
\end{figure}

Among the LLMs with Search Tools, Claude-3.7-Sonnet delivers an impressive performance, with its overall score exceeding Grok Deeper Search, which is perhaps attributable to Claude being allowed to perform multi-turn web searches. Perplexity-Sonar-Reasoning-Pro(high) also demonstrated similar performance. When viewing scores of different models across topics in Figure~\ref{fig:topic_score_heatmap}, we can observe that individual models maintain relatively stable performance across different topic tasks, demonstrating the robustness of our RACE evaluation framework. Interestingly, we discover that for the Chinese task within the transportation topic, all models perform below their respective average levels, indicating that this particular question set presents a higher difficulty level. The generated articles and corresponding scores for all models can be viewed on the leaderboard hosted on Hugging Face Spaces.

\subsubsection{Evaluation on FACT Framework}
Viewing evaluation results by FACT in Table~\ref{tab:main_results}, Deep Research Agents, except Grok, tend to include more Effective Citations than LLMs with Search Tools. Notably, Gemini-2.5-Pro Deep Research achieves an average of 111.21 effective citations in its final reports, significantly outperforming other models. This finding is consistent with its top score in the "Comprehensiveness" dimension of the RACE framework. However, the citation accuracy of Gemini-2.5-Pro Deep Research and OpenAI Deep Research shows a notable gap compared to Perplexity Deep Research, indicating that Perplexity Deep Research possesses a stronger capability to accurately recall relevant content from the retrieved text. Among the LLMs with Search Tools, Claude-3.7-Sonnet achieves the second highest number of Effective Citations and demonstrates strong Citation Accuracy; this also aligns with Claude-3.7-Sonnet w/Search achieving the best overall score in the RACE framework.

\subsection{Human Consistency}
\label{subsec:human_consistency}
Evaluating the quality of deep research reports remains an open-ended task. Therefore, to validate the effectiveness of our proposed RACE framework, we must assess its human consistency. We conduct experiments using 50 Chinese tasks from DeepResearch Bench, with reports generated by four distinct agents. For each task, three domain-expert annotators score these reports. Further details are provided in Appendix~\ref{subsec: human details}.

\subsubsection{Human Data Collection}
\label{subsubsec:human_data_collection}
To gather human judgments, we recruit 70+ annotators with Master's degrees and relevant domain expertise. Using a custom interface, they evaluate reports across four dimensions and overall performance, guided only by basic scoring criteria to minimize bias. Each annotator is limited to a maximum of three queries to ensure diverse perspectives.

\subsubsection{Evaluation Metrics}
\label{subsubsec:metrics}

To validate the consistency between evaluation methods and human judgment, we design four metrics that quantify different aspects of alignment with human evaluations. The detailed calculation processes for all metrics are provided in Appendix~\ref{app:human_consistency_metrics_detailed}.

\paragraph{Pairwise Agreement Rate} This metric measures how often our evaluation method's preferences match human experts' preferences when comparing pairs of reports. It reflects the reliability of our framework in replicating human comparative judgments across all tasks.

\paragraph{Overall Pearson Correlation} This metric quantifies the linear relationship between average model scores from our evaluation method and those from human experts. It demonstrates how well our framework's absolute scoring aligns with human evaluation across all deep research assistant models.

\paragraph{Filtered Average Pearson \& Spearman Correlation}
When computing per-task average correlation coefficients, individual inconsistencies can have a more pronounced impact on the results compared to global correlations. To address this issue, we first filter out tasks where expert judgments show low agreement by removing tasks with negative Intraclass Correlation Coefficients (ICC < 0). The ICC is a statistical measure of rater consistency, and negative values indicate poor inter-rater reliability. After applying this filtering criterion, 37 tasks (out of the original set) remained in our experiment, forming a subset with demonstrably higher expert consensus. We then compute two complementary metrics: the Filtered Average Pearson Correlation (FAP) and the Filtered Average Spearman Correlation (FAS). Together, these filtered metrics provide a more robust assessment of how well automated evaluation aligns with consistent human judgment across different tasks.

\begin{table}[htbp]
\centering
\caption{Comparison of human consistency scores across different evaluation methods. Prefixed with '-', indicating the removal of specific components from the full framework. PAR: Pairwise Agreement Rate; OPC: Overall Pearson Correlation; FAP: Filtered Average Pearson; FAS: Filtered Average Spearman. All metrics are percentages where applicable. Best scores for each metric among automated methods are in \textbf{bold}.}
\label{table:ablation_results}
\small % Use \small or \footnotesize if the table is too wide
\begin{tabular*}{\textwidth}{@{\extracolsep{\fill}}lccccc@{}}
\toprule
Evaluation Method & PAR & OPC & FAP & FAS & Overall Score \\
\midrule
Vanilla Prompt & 58.89 & 98.89 & 40.30 & 43.75 & 60.46 \\
\midrule % Line requested by user
\textbf{RACE(Full)} & \textbf{71.33} & 99.54 & \textbf{60.24} & \textbf{59.12} & \textbf{72.56} \\
\hspace{1em}- No Criteria Weights & 70.67 & 99.62 & 59.83 & 56.27 & 71.60 \\
\hspace{1em}- No Dim Weights & 70.89 & 99.54 & 60.11 & 57.22 & 71.94 \\
\hspace{1em}- No Weights & 71.11 & \textbf{99.69} & 59.46 & 58.17 & 72.11 \\
\hspace{1em}- No Reference & 66.56 & 97.46 & 57.51 & 51.23 & 68.19 \\
Reverse Position & 69.56 & 97.20 & 56.75 & 55.49 & 69.75 \\
Static Criteria & 68.33 & 98.73 & 57.86 & 57.70 & 70.65 \\
\midrule
Human Inter-Agreement & 68.44 & - & - & - & - \\
\bottomrule
\end{tabular*}
\end{table}

\subsubsection{Comparison of Different Evaluation Methods}
\label{subsec:comparison_methods}

Given that existing evaluation methods are generally unsuitable for assessing deep research reports, our experiments focus on comparing the human consistency of several RACE variants against a vanilla prompt baseline. We evaluate the following methods: RACE (Full), Vanilla Prompt (direct scoring by Judge LLM), ablation variants, Reverse Position (swapping reference and target report positions), and Static Criteria (using fixed criteria). As shown in Table~\ref{table:ablation_results}, \textbf{RACE(Full)} exhibits the strongest overall performance, significantly outperforming both the vanilla baseline and other RACE variants. Notably, its Pairwise Agreement Rate also surpasses the inter-agreement rate observed among human experts. These results collectively demonstrate that RACE can reliably and accurately evaluate deep research reports, achieving high human consistency efficiently.

\subsubsection{Comparison of Different Judge LLM}
\label{subsec:comparison_judge_llm} 

Leveraging the RACE framework, we further compare the performance and cost of several leading proprietary LLMs when used as the Judge LLM. As detailed in Table~\ref{table:judge_llm_comparison}, Gemini 2.5 Pro Preview not only achieves the best overall performance but also maintains a competitive average cost (\$0.13 per query), only higher than that of o4-mini. Therefore, to effectively balance performance and cost, we select Gemini 2.5 Pro Preview as the Judge LLM in our final framework.

\begin{table}[htbp]
\centering
\caption{Comparison of human consistency scores and average cost using different Judge LLMs within the RACE(Full) framework. PAR: Pairwise Agreement Rate; OPC: Overall Pearson Correlation; FAP: Filtered Average Pearson; FAS: Filtered Average Spearman; Overall: Overall Score. Consistency metrics are percentages. The best for each metric are highlighted in \textbf{bold}}
\label{table:judge_llm_comparison}
\small % Use \small or \footnotesize if the table is too wide
\begin{tabular*}{\textwidth}{@{\extracolsep{\fill}}lcccccc@{}}
\toprule
Judge LLM & PAR & OPC & FAP & FAS & Overall & Avg. Cost (\$) \\
\midrule
Gemini 2.5 Pro Preview & \textbf{71.33} & \textbf{99.54} & \textbf{60.24} & 59.12 & \textbf{72.56} & 0.13 \\
o3 & 68.11 & 96.22 & 57.64 & 52.36 & 68.58 & 0.37 \\
o4-mini & 70.89 & 97.06 & 59.54 & 59.02 & 71.63 & \textbf{0.04} \\
Claude 3.7 Sonnet & 70.78 & 96.53 & 58.22 & \textbf{63.61} & 72.28 & 0.47 \\
\bottomrule
\end{tabular*}
\end{table}

\section{Related Work}

\paragraph{LLM-based Agent Evaluation}
With the comprehensive advancement of LLM capabilities, LLM-based Agents are increasingly being applied to real-world scenarios \cite{mon-williams_embodied_2025, wang_survey_2025}, promising to significantly alter many aspects of daily life and professional work. Yao's blog \cite{yao_second_nodate} emphasizes that defining more realistic problems and designing novel evaluation methods are crucial for developing more practical AI Agent systems. Numerous evaluations have already been designed specifically for Agents, targeting diverse capabilities. These include evaluations for agents in scientific domains \cite{chan_mle-bench_2025, laurent_lab-bench_2024, mitchener_bixbench_2025, chen_scienceagentbench_2025}, creative writing \cite{wu_writingbench_2025, bai_longwriter_2024, que_hellobench_2024}, code generation and software engineering \cite{jimenez_swe-bench_2024, zhuo_bigcodebench_2025, quan_codeelo_2025, jain_livecodebench_2024, xiao_csr-bench_2025}, and in their roles as human assistants, often enhanced by capabilities such as web Browse and tool-use \cite{wei_browsecomp_2025, zhou_browsecomp-zh_2025, yan_berkeley_2024, deng_mobile-bench_2024, wang_mobileagentbench_2024}. This perspective underscores our belief that constructing benchmarks specifically designed for Deep Research Agent, grounded in real-world scenarios, alongside developing human-aligned evaluation methods, is urgently needed to guide the development of AI agent systems.

\paragraph{Deep Research Agent}
Following the release of Deep Research Agents (DRAs) by OpenAI \cite{openai_introducing_nodate} and Google's Gemini \cite{google_deep_2025}, these agents gained significant attention and became one of the most widely deployed categories of LLM-based agents. Subsequently, related works \cite{langchain-ai_langchain-aiopen_deep_research_nodate, li_webthinker_2025, zheng_deepresearcher_2025} quickly followed, introducing their own designed DRA frameworks. However, the field still lacks a standardized evaluation methodology for these DRAs, preventing meaningful comparative analysis of their capabilities. Among these works, some utilize QA datasets \cite{phan_humanitys_2025, mialon_gaia_2023, wu_webwalker_2025} as evaluation metrics; however, this approach neither aligns with real-world DRA applications nor comprehensively assesses their broader capabilities. Others employ the LLM-as-a-judge methodology \cite{zheng_judging_2023}, yet these efforts lack both a comprehensive framework design and verification of human consistency. In contrast, our DeepResearch Bench addresses this gap by providing a systematic and unified evaluation method with strong human consistency, thereby supporting subsequent DRA development and assessment.

\section{Conclusion}
In this work, we introduce DeepResearch Bench, the first comprehensive benchmark for evaluating the report generation and web retrieval capabilities of Deep Research Agents. Comprising 100 high-quality research tasks across 22 distinct domains, this benchmark is meticulously curated to reflect authentic user needs. Our key evaluation frameworks, RACE and FACT, have demonstrated high consistency with human judgments, affirming their reliability. We hope DeepResearch Bench will guide developers and researchers in constructing more powerful and human-centric AI agent systems that truly address genuine user requirements.

\newpage

\medskip

\bibliographystyle{plain} 
\bibliography{references}

\begin{thebibliography}{10}

\bibitem{bai_longwriter_2024}
Yushi Bai, Jiajie Zhang, Xin Lv, Linzhi Zheng, Siqi Zhu, Lei Hou, Yuxiao Dong, Jie Tang, and Juanzi Li.
\newblock {LongWriter}: {Unleashing} 10,000+ {Word} {Generation} from {Long} {Context} {LLMs}, August 2024.
\newblock arXiv:2408.07055 [cs].

\bibitem{chan_mle-bench_2025}
Jun~Shern Chan, Neil Chowdhury, Oliver Jaffe, James Aung, Dane Sherburn, Evan Mays, Giulio Starace, Kevin Liu, Leon Maksin, Tejal Patwardhan, Lilian Weng, and Aleksander Mądry.
\newblock {MLE}-bench: {Evaluating} {Machine} {Learning} {Agents} on {Machine} {Learning} {Engineering}, February 2025.
\newblock arXiv:2410.07095 [cs].

\bibitem{chen_scienceagentbench_2025}
Ziru Chen, Shijie Chen, Yuting Ning, Qianheng Zhang, Boshi Wang, Botao Yu, Yifei Li, Zeyi Liao, Chen Wei, Zitong Lu, Vishal Dey, Mingyi Xue, Frazier~N. Baker, Benjamin Burns, Daniel Adu-Ampratwum, Xuhui Huang, Xia Ning, Song Gao, Yu~Su, and Huan Sun.
\newblock {ScienceAgentBench}: {Toward} {Rigorous} {Assessment} of {Language} {Agents} for {Data}-{Driven} {Scientific} {Discovery}, March 2025.
\newblock arXiv:2410.05080 [cs].

\bibitem{deepseek-ai_deepseek-aideepseek-v3-0324_2025}
DeepSeek-AI.
\newblock deepseek-ai/{DeepSeek}-{V3}-0324 · {Hugging} {Face}, March 2025.

\bibitem{deepseek-ai_deepseek-r1_2025}
DeepSeek-AI, Daya Guo, Dejian Yang, Haowei Zhang, Junxiao Song, Ruoyu Zhang, Runxin Xu, Qihao Zhu, Shirong Ma, Peiyi Wang, Xiao Bi, Xiaokang Zhang, Xingkai Yu, Yu~Wu, Z.~F. Wu, Zhibin Gou, Zhihong Shao, Zhuoshu Li, Ziyi Gao, Aixin Liu, Bing Xue, Bingxuan Wang, Bochao Wu, Bei Feng, Chengda Lu, Chenggang Zhao, Chengqi Deng, Chenyu Zhang, Chong Ruan, Damai Dai, Deli Chen, Dongjie Ji, Erhang Li, Fangyun Lin, Fucong Dai, Fuli Luo, Guangbo Hao, Guanting Chen, Guowei Li, H.~Zhang, Han Bao, Hanwei Xu, Haocheng Wang, Honghui Ding, Huajian Xin, Huazuo Gao, Hui Qu, Hui Li, Jianzhong Guo, Jiashi Li, Jiawei Wang, Jingchang Chen, Jingyang Yuan, Junjie Qiu, Junlong Li, J.~L. Cai, Jiaqi Ni, Jian Liang, Jin Chen, Kai Dong, Kai Hu, Kaige Gao, Kang Guan, Kexin Huang, Kuai Yu, Lean Wang, Lecong Zhang, Liang Zhao, Litong Wang, Liyue Zhang, Lei Xu, Leyi Xia, Mingchuan Zhang, Minghua Zhang, Minghui Tang, Meng Li, Miaojun Wang, Mingming Li, Ning Tian, Panpan Huang, Peng Zhang, Qiancheng Wang, Qinyu Chen, Qiushi Du, Ruiqi Ge, Ruisong
  Zhang, Ruizhe Pan, Runji Wang, R.~J. Chen, R.~L. Jin, Ruyi Chen, Shanghao Lu, Shangyan Zhou, Shanhuang Chen, Shengfeng Ye, Shiyu Wang, Shuiping Yu, Shunfeng Zhou, Shuting Pan, S.~S. Li, Shuang Zhou, Shaoqing Wu, Shengfeng Ye, Tao Yun, Tian Pei, Tianyu Sun, T.~Wang, Wangding Zeng, Wanjia Zhao, Wen Liu, Wenfeng Liang, Wenjun Gao, Wenqin Yu, Wentao Zhang, W.~L. Xiao, Wei An, Xiaodong Liu, Xiaohan Wang, Xiaokang Chen, Xiaotao Nie, Xin Cheng, Xin Liu, Xin Xie, Xingchao Liu, Xinyu Yang, Xinyuan Li, Xuecheng Su, Xuheng Lin, X.~Q. Li, Xiangyue Jin, Xiaojin Shen, Xiaosha Chen, Xiaowen Sun, Xiaoxiang Wang, Xinnan Song, Xinyi Zhou, Xianzu Wang, Xinxia Shan, Y.~K. Li, Y.~Q. Wang, Y.~X. Wei, Yang Zhang, Yanhong Xu, Yao Li, Yao Zhao, Yaofeng Sun, Yaohui Wang, Yi~Yu, Yichao Zhang, Yifan Shi, Yiliang Xiong, Ying He, Yishi Piao, Yisong Wang, Yixuan Tan, Yiyang Ma, Yiyuan Liu, Yongqiang Guo, Yuan Ou, Yuduan Wang, Yue Gong, Yuheng Zou, Yujia He, Yunfan Xiong, Yuxiang Luo, Yuxiang You, Yuxuan Liu, Yuyang Zhou, Y.~X. Zhu,
  Yanhong Xu, Yanping Huang, Yaohui Li, Yi~Zheng, Yuchen Zhu, Yunxian Ma, Ying Tang, Yukun Zha, Yuting Yan, Z.~Z. Ren, Zehui Ren, Zhangli Sha, Zhe Fu, Zhean Xu, Zhenda Xie, Zhengyan Zhang, Zhewen Hao, Zhicheng Ma, Zhigang Yan, Zhiyu Wu, Zihui Gu, Zijia Zhu, Zijun Liu, Zilin Li, Ziwei Xie, Ziyang Song, Zizheng Pan, Zhen Huang, Zhipeng Xu, Zhongyu Zhang, and Zhen Zhang.
\newblock {DeepSeek}-{R1}: {Incentivizing} {Reasoning} {Capability} in {LLMs} via {Reinforcement} {Learning}, January 2025.
\newblock arXiv:2501.12948 [cs].

\bibitem{deng_mobile-bench_2024}
Shihan Deng, Weikai Xu, Hongda Sun, Wei Liu, Tao Tan, Jianfeng Liu, Ang Li, Jian Luan, Bin Wang, Rui Yan, and Shuo Shang.
\newblock Mobile-{Bench}: {An} {Evaluation} {Benchmark} for {LLM}-based {Mobile} {Agents}, July 2024.
\newblock arXiv:2407.00993 [cs].

\bibitem{google_deep_2025}
Gemini Google.
\newblock Deep {Research} is now available on {Gemini} 2.5 {Pro} {Experimental}., April 2025.

\bibitem{hong_metagpt_2024}
Sirui Hong, Mingchen Zhuge, Jiaqi Chen, Xiawu Zheng, Yuheng Cheng, Ceyao Zhang, Jinlin Wang, Zili Wang, Steven Ka~Shing Yau, Zijuan Lin, Liyang Zhou, Chenyu Ran, Lingfeng Xiao, Chenglin Wu, and Jürgen Schmidhuber.
\newblock {MetaGPT}: {Meta} {Programming} for {A} {Multi}-{Agent} {Collaborative} {Framework}, November 2024.
\newblock arXiv:2308.00352 [cs].

\bibitem{jain_livecodebench_2024}
Naman Jain, King Han, Alex Gu, Wen-Ding Li, Fanjia Yan, Tianjun Zhang, Sida Wang, Armando Solar-Lezama, Koushik Sen, and Ion Stoica.
\newblock {LiveCodeBench}: {Holistic} and {Contamination} {Free} {Evaluation} of {Large} {Language} {Models} for {Code}, June 2024.
\newblock arXiv:2403.07974 [cs].

\bibitem{jimenez_swe-bench_2024}
Carlos~E. Jimenez, John Yang, Alexander Wettig, Shunyu Yao, Kexin Pei, Ofir Press, and Karthik Narasimhan.
\newblock {SWE}-bench: {Can} {Language} {Models} {Resolve} {Real}-{World} {GitHub} {Issues}?, November 2024.
\newblock arXiv:2310.06770 [cs].

\bibitem{langchain-ai_langchain-aiopen_deep_research_nodate}
langchain ai.
\newblock langchain-ai/open\_deep\_research.

\bibitem{laurent_lab-bench_2024}
Jon~M. Laurent, Joseph~D. Janizek, Michael Ruzo, Michaela~M. Hinks, Michael~J. Hammerling, Siddharth Narayanan, Manvitha Ponnapati, Andrew~D. White, and Samuel~G. Rodriques.
\newblock {LAB}-{Bench}: {Measuring} {Capabilities} of {Language} {Models} for {Biology} {Research}, July 2024.
\newblock arXiv:2407.10362 [cs].

\bibitem{li_webthinker_2025}
Xiaoxi Li, Jiajie Jin, Guanting Dong, Hongjin Qian, Yutao Zhu, Yongkang Wu, Ji-Rong Wen, and Zhicheng Dou.
\newblock {WebThinker}: {Empowering} {Large} {Reasoning} {Models} with {Deep} {Research} {Capability}, April 2025.
\newblock arXiv:2504.21776 [cs].

\bibitem{liu_agentbench_2023}
Xiao Liu, Hao Yu, Hanchen Zhang, Yifan Xu, Xuanyu Lei, Hanyu Lai, Yu~Gu, Hangliang Ding, Kaiwen Men, Kejuan Yang, Shudan Zhang, Xiang Deng, Aohan Zeng, Zhengxiao Du, Chenhui Zhang, Sheng Shen, Tianjun Zhang, Yu~Su, Huan Sun, Minlie Huang, Yuxiao Dong, and Jie Tang.
\newblock {AgentBench}: {Evaluating} {LLMs} as {Agents}, October 2023.
\newblock arXiv:2308.03688 [cs].

\bibitem{masterman_landscape_2024}
Tula Masterman, Sandi Besen, Mason Sawtell, and Alex Chao.
\newblock The {Landscape} of {Emerging} {AI} {Agent} {Architectures} for {Reasoning}, {Planning}, and {Tool} {Calling}: {A} {Survey}, April 2024.
\newblock arXiv:2404.11584 [cs].

\bibitem{mialon_gaia_2023}
Grégoire Mialon, Clémentine Fourrier, Craig Swift, Thomas Wolf, Yann LeCun, and Thomas Scialom.
\newblock {GAIA}: a benchmark for {General} {AI} {Assistants}, November 2023.
\newblock arXiv:2311.12983 [cs].

\bibitem{mitchener_bixbench_2025}
Ludovico Mitchener, Jon~M. Laurent, Benjamin Tenmann, Siddharth Narayanan, Geemi~P. Wellawatte, Andrew White, Lorenzo Sani, and Samuel~G. Rodriques.
\newblock {BixBench}: a {Comprehensive} {Benchmark} for {LLM}-based {Agents} in {Computational} {Biology}, March 2025.
\newblock arXiv:2503.00096 [q-bio].

\bibitem{mon-williams_embodied_2025}
Ruaridh Mon-Williams, Gen Li, Ran Long, Wenqian Du, and Christopher~G. Lucas.
\newblock Embodied large language models enable robots to complete complex tasks in unpredictable environments.
\newblock {\em Nature Machine Intelligence}, 7(4):592--601, April 2025.
\newblock Publisher: Nature Publishing Group.

\bibitem{openai_introducing_nodate}
OpenAI.
\newblock Introducing deep research {\textbar} {OpenAI}.

\bibitem{openai_openai_2024}
OpenAI, Aaron Jaech, Adam Kalai, Adam Lerer, Adam Richardson, Ahmed El-Kishky, Aiden Low, Alec Helyar, Aleksander Madry, Alex Beutel, Alex Carney, Alex Iftimie, Alex Karpenko, Alex~Tachard Passos, Alexander Neitz, Alexander Prokofiev, Alexander Wei, Allison Tam, Ally Bennett, Ananya Kumar, Andre Saraiva, Andrea Vallone, Andrew Duberstein, Andrew Kondrich, Andrey Mishchenko, Andy Applebaum, Angela Jiang, Ashvin Nair, Barret Zoph, Behrooz Ghorbani, Ben Rossen, Benjamin Sokolowsky, Boaz Barak, Bob McGrew, Borys Minaiev, Botao Hao, Bowen Baker, Brandon Houghton, Brandon McKinzie, Brydon Eastman, Camillo Lugaresi, Cary Bassin, Cary Hudson, Chak~Ming Li, Charles~de Bourcy, Chelsea Voss, Chen Shen, Chong Zhang, Chris Koch, Chris Orsinger, Christopher Hesse, Claudia Fischer, Clive Chan, Dan Roberts, Daniel Kappler, Daniel Levy, Daniel Selsam, David Dohan, David Farhi, David Mely, David Robinson, Dimitris Tsipras, Doug Li, Dragos Oprica, Eben Freeman, Eddie Zhang, Edmund Wong, Elizabeth Proehl, Enoch Cheung, Eric
  Mitchell, Eric Wallace, Erik Ritter, Evan Mays, Fan Wang, Felipe~Petroski Such, Filippo Raso, Florencia Leoni, Foivos Tsimpourlas, Francis Song, Fred~von Lohmann, Freddie Sulit, Geoff Salmon, Giambattista Parascandolo, Gildas Chabot, Grace Zhao, Greg Brockman, Guillaume Leclerc, Hadi Salman, Haiming Bao, Hao Sheng, Hart Andrin, Hessam Bagherinezhad, Hongyu Ren, Hunter Lightman, Hyung~Won Chung, Ian Kivlichan, Ian O'Connell, Ian Osband, Ignasi~Clavera Gilaberte, Ilge Akkaya, Ilya Kostrikov, Ilya Sutskever, Irina Kofman, Jakub Pachocki, James Lennon, Jason Wei, Jean Harb, Jerry Twore, Jiacheng Feng, Jiahui Yu, Jiayi Weng, Jie Tang, Jieqi Yu, Joaquin~Quiñonero Candela, Joe Palermo, Joel Parish, Johannes Heidecke, John Hallman, John Rizzo, Jonathan Gordon, Jonathan Uesato, Jonathan Ward, Joost Huizinga, Julie Wang, Kai Chen, Kai Xiao, Karan Singhal, Karina Nguyen, Karl Cobbe, Katy Shi, Kayla Wood, Kendra Rimbach, Keren Gu-Lemberg, Kevin Liu, Kevin Lu, Kevin Stone, Kevin Yu, Lama Ahmad, Lauren Yang, Leo Liu,
  Leon Maksin, Leyton Ho, Liam Fedus, Lilian Weng, Linden Li, Lindsay McCallum, Lindsey Held, Lorenz Kuhn, Lukas Kondraciuk, Lukasz Kaiser, Luke Metz, Madelaine Boyd, Maja Trebacz, Manas Joglekar, Mark Chen, Marko Tintor, Mason Meyer, Matt Jones, Matt Kaufer, Max Schwarzer, Meghan Shah, Mehmet Yatbaz, Melody~Y. Guan, Mengyuan Xu, Mengyuan Yan, Mia Glaese, Mianna Chen, Michael Lampe, Michael Malek, Michele Wang, Michelle Fradin, Mike McClay, Mikhail Pavlov, Miles Wang, Mingxuan Wang, Mira Murati, Mo~Bavarian, Mostafa Rohaninejad, Nat McAleese, Neil Chowdhury, Neil Chowdhury, Nick Ryder, Nikolas Tezak, Noam Brown, Ofir Nachum, Oleg Boiko, Oleg Murk, Olivia Watkins, Patrick Chao, Paul Ashbourne, Pavel Izmailov, Peter Zhokhov, Rachel Dias, Rahul Arora, Randall Lin, Rapha~Gontijo Lopes, Raz Gaon, Reah Miyara, Reimar Leike, Renny Hwang, Rhythm Garg, Robin Brown, Roshan James, Rui Shu, Ryan Cheu, Ryan Greene, Saachi Jain, Sam Altman, Sam Toizer, Sam Toyer, Samuel Miserendino, Sandhini Agarwal, Santiago Hernandez,
  Sasha Baker, Scott McKinney, Scottie Yan, Shengjia Zhao, Shengli Hu, Shibani Santurkar, Shraman~Ray Chaudhuri, Shuyuan Zhang, Siyuan Fu, Spencer Papay, Steph Lin, Suchir Balaji, Suvansh Sanjeev, Szymon Sidor, Tal Broda, Aidan Clark, Tao Wang, Taylor Gordon, Ted Sanders, Tejal Patwardhan, Thibault Sottiaux, Thomas Degry, Thomas Dimson, Tianhao Zheng, Timur Garipov, Tom Stasi, Trapit Bansal, Trevor Creech, Troy Peterson, Tyna Eloundou, Valerie Qi, Vineet Kosaraju, Vinnie Monaco, Vitchyr Pong, Vlad Fomenko, Weiyi Zheng, Wenda Zhou, Wes McCabe, Wojciech Zaremba, Yann Dubois, Yinghai Lu, Yining Chen, Young Cha, Yu~Bai, Yuchen He, Yuchen Zhang, Yunyun Wang, Zheng Shao, and Zhuohan Li.
\newblock {OpenAI} o1 {System} {Card}, December 2024.
\newblock arXiv:2412.16720 [cs].

\bibitem{phan_humanitys_2025}
Long Phan, Alice Gatti, Ziwen Han, Nathaniel Li, Josephina Hu, Hugh Zhang, Chen Bo~Calvin Zhang, Mohamed Shaaban, John Ling, Sean Shi, Michael Choi, Anish Agrawal, Arnav Chopra, Adam Khoja, Ryan Kim, Richard Ren, Jason Hausenloy, Oliver Zhang, Mantas Mazeika, Dmitry Dodonov, Tung Nguyen, Jaeho Lee, Daron Anderson, Mikhail Doroshenko, Alun~Cennyth Stokes, Mobeen Mahmood, Oleksandr Pokutnyi, Oleg Iskra, Jessica~P. Wang, John-Clark Levin, Mstyslav Kazakov, Fiona Feng, Steven~Y. Feng, Haoran Zhao, Michael Yu, Varun Gangal, Chelsea Zou, Zihan Wang, Serguei Popov, Robert Gerbicz, Geoff Galgon, Johannes Schmitt, Will Yeadon, Yongki Lee, Scott Sauers, Alvaro Sanchez, Fabian Giska, Marc Roth, Søren Riis, Saiteja Utpala, Noah Burns, Gashaw~M. Goshu, Mohinder~Maheshbhai Naiya, Chidozie Agu, Zachary Giboney, Antrell Cheatom, Francesco Fournier-Facio, Sarah-Jane Crowson, Lennart Finke, Zerui Cheng, Jennifer Zampese, Ryan~G. Hoerr, Mark Nandor, Hyunwoo Park, Tim Gehrunger, Jiaqi Cai, Ben McCarty, Alexis~C. Garretson,
  Edwin Taylor, Damien Sileo, Qiuyu Ren, Usman Qazi, Lianghui Li, Jungbae Nam, John~B. Wydallis, Pavel Arkhipov, Jack Wei~Lun Shi, Aras Bacho, Chris~G. Willcocks, Hangrui Cao, Sumeet Motwani, Emily de~Oliveira Santos, Johannes Veith, Edward Vendrow, Doru Cojoc, Kengo Zenitani, Joshua Robinson, Longke Tang, Yuqi Li, Joshua Vendrow, Natanael~Wildner Fraga, Vladyslav Kuchkin, Andrey~Pupasov Maksimov, Pierre Marion, Denis Efremov, Jayson Lynch, Kaiqu Liang, Aleksandar Mikov, Andrew Gritsevskiy, Julien Guillod, Gözdenur Demir, Dakotah Martinez, Ben Pageler, Kevin Zhou, Saeed Soori, Ori Press, Henry Tang, Paolo Rissone, Sean~R. Green, Lina Brüssel, Moon Twayana, Aymeric Dieuleveut, Joseph~Marvin Imperial, Ameya Prabhu, Jinzhou Yang, Nick Crispino, Arun Rao, Dimitri Zvonkine, Gabriel Loiseau, Mikhail Kalinin, Marco Lukas, Ciprian Manolescu, Nate Stambaugh, Subrata Mishra, Tad Hogg, Carlo Bosio, Brian~P. Coppola, Julian Salazar, Jaehyeok Jin, Rafael Sayous, Stefan Ivanov, Philippe Schwaller, Shaipranesh
  Senthilkuma, Andres~M. Bran, Andres Algaba, Kelsey Van~den Houte, Lynn Van~Der Sypt, Brecht Verbeken, David Noever, Alexei Kopylov, Benjamin Myklebust, Bikun Li, Lisa Schut, Evgenii Zheltonozhskii, Qiaochu Yuan, Derek Lim, Richard Stanley, Tong Yang, John Maar, Julian Wykowski, Martí Oller, Anmol Sahu, Cesare~Giulio Ardito, Yuzheng Hu, Ariel Ghislain~Kemogne Kamdoum, Alvin Jin, Tobias~Garcia Vilchis, Yuexuan Zu, Martin Lackner, James Koppel, Gongbo Sun, Daniil~S. Antonenko, Steffi Chern, Bingchen Zhao, Pierrot Arsene, Joseph~M. Cavanagh, Daofeng Li, Jiawei Shen, Donato Crisostomi, Wenjin Zhang, Ali Dehghan, Sergey Ivanov, David Perrella, Nurdin Kaparov, Allen Zang, Ilia Sucholutsky, Arina Kharlamova, Daniil Orel, Vladislav Poritski, Shalev Ben-David, Zachary Berger, Parker Whitfill, Michael Foster, Daniel Munro, Linh Ho, Shankar Sivarajan, Dan~Bar Hava, Aleksey Kuchkin, David Holmes, Alexandra Rodriguez-Romero, Frank Sommerhage, Anji Zhang, Richard Moat, Keith Schneider, Zakayo Kazibwe, Don Clarke,
  Dae~Hyun Kim, Felipe~Meneguitti Dias, Sara Fish, Veit Elser, Tobias Kreiman, Victor Efren~Guadarrama Vilchis, Immo Klose, Ujjwala Anantheswaran, Adam Zweiger, Kaivalya Rawal, Jeffery Li, Jeremy Nguyen, Nicolas Daans, Haline Heidinger, Maksim Radionov, Václav Rozhoň, Vincent Ginis, Christian Stump, Niv Cohen, Rafał Poświata, Josef Tkadlec, Alan Goldfarb, Chenguang Wang, Piotr Padlewski, Stanislaw Barzowski, Kyle Montgomery, Ryan Stendall, Jamie Tucker-Foltz, Jack Stade, T.~Ryan Rogers, Tom Goertzen, Declan Grabb, Abhishek Shukla, Alan Givré, John~Arnold Ambay, Archan Sen, Muhammad~Fayez Aziz, Mark~H. Inlow, Hao He, Ling Zhang, Younesse Kaddar, Ivar Ängquist, Yanxu Chen, Harrison~K. Wang, Kalyan Ramakrishnan, Elliott Thornley, Antonio Terpin, Hailey Schoelkopf, Eric Zheng, Avishy Carmi, Ethan D.~L. Brown, Kelin Zhu, Max Bartolo, Richard Wheeler, Martin Stehberger, Peter Bradshaw, J.~P. Heimonen, Kaustubh Sridhar, Ido Akov, Jennifer Sandlin, Yury Makarychev, Joanna Tam, Hieu Hoang, David~M. Cunningham,
  Vladimir Goryachev, Demosthenes Patramanis, Michael Krause, Andrew Redenti, David Aldous, Jesyin Lai, Shannon Coleman, Jiangnan Xu, Sangwon Lee, Ilias Magoulas, Sandy Zhao, Ning Tang, Michael~K. Cohen, Orr Paradise, Jan~Hendrik Kirchner, Maksym Ovchynnikov, Jason~O. Matos, Adithya Shenoy, Michael Wang, Yuzhou Nie, Anna Sztyber-Betley, Paolo Faraboschi, Robin Riblet, Jonathan Crozier, Shiv Halasyamani, Shreyas Verma, Prashant Joshi, Eli Meril, Ziqiao Ma, Jérémy Andréoletti, Raghav Singhal, Jacob Platnick, Volodymyr Nevirkovets, Luke Basler, Alexander Ivanov, Seri Khoury, Nils Gustafsson, Marco Piccardo, Hamid Mostaghimi, Qijia Chen, Virendra Singh, Tran~Quoc Khánh, Paul Rosu, Hannah Szlyk, Zachary Brown, Himanshu Narayan, Aline Menezes, Jonathan Roberts, William Alley, Kunyang Sun, Arkil Patel, Max Lamparth, Anka Reuel, Linwei Xin, Hanmeng Xu, Jacob Loader, Freddie Martin, Zixuan Wang, Andrea Achilleos, Thomas Preu, Tomek Korbak, Ida Bosio, Fereshteh Kazemi, Ziye Chen, Biró Bálint, Eve J.~Y. Lo, Jiaqi
  Wang, Maria Inês~S. Nunes, Jeremiah Milbauer, M.~Saiful Bari, Zihao Wang, Behzad Ansarinejad, Yewen Sun, Stephane Durand, Hossam Elgnainy, Guillaume Douville, Daniel Tordera, George Balabanian, Hew Wolff, Lynna Kvistad, Hsiaoyun Milliron, Ahmad Sakor, Murat Eron, Andrew Favre~D. O, Shailesh Shah, Xiaoxiang Zhou, Firuz Kamalov, Sherwin Abdoli, Tim Santens, Shaul Barkan, Allison Tee, Robin Zhang, Alessandro Tomasiello, G.~Bruno~De Luca, Shi-Zhuo Looi, Vinh-Kha Le, Noam Kolt, Jiayi Pan, Emma Rodman, Jacob Drori, Carl~J. Fossum, Niklas Muennighoff, Milind Jagota, Ronak Pradeep, Honglu Fan, Jonathan Eicher, Michael Chen, Kushal Thaman, William Merrill, Moritz Firsching, Carter Harris, Stefan Ciobâcă, Jason Gross, Rohan Pandey, Ilya Gusev, Adam Jones, Shashank Agnihotri, Pavel Zhelnov, Mohammadreza Mofayezi, Alexander Piperski, David~K. Zhang, Kostiantyn Dobarskyi, Roman Leventov, Ignat Soroko, Joshua Duersch, Vage Taamazyan, Andrew Ho, Wenjie Ma, William Held, Ruicheng Xian, Armel~Randy Zebaze, Mohanad
  Mohamed, Julian~Noah Leser, Michelle~X. Yuan, Laila Yacar, Johannes Lengler, Katarzyna Olszewska, Claudio~Di Fratta, Edson Oliveira, Joseph~W. Jackson, Andy Zou, Muthu Chidambaram, Timothy Manik, Hector Haffenden, Dashiell Stander, Ali Dasouqi, Alexander Shen, Bita Golshani, David Stap, Egor Kretov, Mikalai Uzhou, Alina~Borisovna Zhidkovskaya, Nick Winter, Miguel~Orbegozo Rodriguez, Robert Lauff, Dustin Wehr, Colin Tang, Zaki Hossain, Shaun Phillips, Fortuna Samuele, Fredrik Ekström, Angela Hammon, Oam Patel, Faraz Farhidi, George Medley, Forough Mohammadzadeh, Madellene Peñaflor, Haile Kassahun, Alena Friedrich, Rayner~Hernandez Perez, Daniel Pyda, Taom Sakal, Omkar Dhamane, Ali~Khajegili Mirabadi, Eric Hallman, Kenchi Okutsu, Mike Battaglia, Mohammad Maghsoudimehrabani, Alon Amit, Dave Hulbert, Roberto Pereira, Simon Weber, Handoko, Anton Peristyy, Stephen Malina, Mustafa Mehkary, Rami Aly, Frank Reidegeld, Anna-Katharina Dick, Cary Friday, Mukhwinder Singh, Hassan Shapourian, Wanyoung Kim, Mariana
  Costa, Hubeyb Gurdogan, Harsh Kumar, Chiara Ceconello, Chao Zhuang, Haon Park, Micah Carroll, Andrew~R. Tawfeek, Stefan Steinerberger, Daattavya Aggarwal, Michael Kirchhof, Linjie Dai, Evan Kim, Johan Ferret, Jainam Shah, Yuzhou Wang, Minghao Yan, Krzysztof Burdzy, Lixin Zhang, Antonio Franca, Diana~T. Pham, Kang~Yong Loh, Joshua Robinson, Abram Jackson, Paolo Giordano, Philipp Petersen, Adrian Cosma, Jesus Colino, Colin White, Jacob Votava, Vladimir Vinnikov, Ethan Delaney, Petr Spelda, Vit Stritecky, Syed~M. Shahid, Jean-Christophe Mourrat, Lavr Vetoshkin, Koen Sponselee, Renas Bacho, Zheng-Xin Yong, Florencia de~la Rosa, Nathan Cho, Xiuyu Li, Guillaume Malod, Orion Weller, Guglielmo Albani, Leon Lang, Julien Laurendeau, Dmitry Kazakov, Fatimah Adesanya, Julien Portier, Lawrence Hollom, Victor Souza, Yuchen~Anna Zhou, Julien Degorre, Yiğit Yalın, Gbenga~Daniel Obikoya, Rai, Filippo Bigi, M.~C. Boscá, Oleg Shumar, Kaniuar Bacho, Gabriel Recchia, Mara Popescu, Nikita Shulga, Ngefor~Mildred Tanwie, Thomas
  C.~H. Lux, Ben Rank, Colin Ni, Matthew Brooks, Alesia Yakimchyk, Huanxu, Liu, Stefano Cavalleri, Olle Häggström, Emil Verkama, Joshua Newbould, Hans Gundlach, Leonor Brito-Santana, Brian Amaro, Vivek Vajipey, Rynaa Grover, Ting Wang, Yosi Kratish, Wen-Ding Li, Sivakanth Gopi, Andrea Caciolai, Christian Schroeder~de Witt, Pablo Hernández-Cámara, Emanuele Rodolà, Jules Robins, Dominic Williamson, Vincent Cheng, Brad Raynor, Hao Qi, Ben Segev, Jingxuan Fan, Sarah Martinson, Erik~Y. Wang, Kaylie Hausknecht, Michael~P. Brenner, Mao Mao, Christoph Demian, Peyman Kassani, Xinyu Zhang, David Avagian, Eshawn~Jessica Scipio, Alon Ragoler, Justin Tan, Blake Sims, Rebeka Plecnik, Aaron Kirtland, Omer~Faruk Bodur, D.~P. Shinde, Yan Carlos~Leyva Labrador, Zahra Adoul, Mohamed Zekry, Ali Karakoc, Tania C.~B. Santos, Samir Shamseldeen, Loukmane Karim, Anna Liakhovitskaia, Nate Resman, Nicholas Farina, Juan~Carlos Gonzalez, Gabe Maayan, Earth Anderson, Rodrigo De~Oliveira Pena, Elizabeth Kelley, Hodjat Mariji, Rasoul
  Pouriamanesh, Wentao Wu, Ross Finocchio, Ismail Alarab, Joshua Cole, Danyelle Ferreira, Bryan Johnson, Mohammad Safdari, Liangti Dai, Siriphan Arthornthurasuk, Isaac~C. McAlister, Alejandro~José Moyano, Alexey Pronin, Jing Fan, Angel Ramirez-Trinidad, Yana Malysheva, Daphiny Pottmaier, Omid Taheri, Stanley Stepanic, Samuel Perry, Luke Askew, Raúl Adrián~Huerta Rodríguez, Ali M.~R. Minissi, Ricardo Lorena, Krishnamurthy Iyer, Arshad~Anil Fasiludeen, Ronald Clark, Josh Ducey, Matheus Piza, Maja Somrak, Eric Vergo, Juehang Qin, Benjámin Borbás, Eric Chu, Jack Lindsey, Antoine Jallon, I.~M.~J. McInnis, Evan Chen, Avi Semler, Luk Gloor, Tej Shah, Marc Carauleanu, Pascal Lauer, Tran~Đuc Huy, Hossein Shahrtash, Emilien Duc, Lukas Lewark, Assaf Brown, Samuel Albanie, Brian Weber, Warren~S. Vaz, Pierre Clavier, Yiyang Fan, Gabriel Poesia Reis~e Silva, Long, Lian, Marcus Abramovitch, Xi~Jiang, Sandra Mendoza, Murat Islam, Juan Gonzalez, Vasilios Mavroudis, Justin Xu, Pawan Kumar, Laxman~Prasad Goswami, Daniel
  Bugas, Nasser Heydari, Ferenc Jeanplong, Thorben Jansen, Antonella Pinto, Archimedes Apronti, Abdallah Galal, Ng~Ze-An, Ankit Singh, Tong Jiang, Joan of~Arc Xavier, Kanu~Priya Agarwal, Mohammed Berkani, Gang Zhang, Zhehang Du, Benedito Alves de~Oliveira Junior, Dmitry Malishev, Nicolas Remy, Taylor~D. Hartman, Tim Tarver, Stephen Mensah, Gautier~Abou Loume, Wiktor Morak, Farzad Habibi, Sarah Hoback, Will Cai, Javier Gimenez, Roselynn~Grace Montecillo, Jakub Łucki, Russell Campbell, Asankhaya Sharma, Khalida Meer, Shreen Gul, Daniel~Espinosa Gonzalez, Xavier Alapont, Alex Hoover, Gunjan Chhablani, Freddie Vargus, Arunim Agarwal, Yibo Jiang, Deepakkumar Patil, David Outevsky, Kevin~Joseph Scaria, Rajat Maheshwari, Abdelkader Dendane, Priti Shukla, Ashley Cartwright, Sergei Bogdanov, Niels Mündler, Sören Möller, Luca Arnaboldi, Kunvar Thaman, Muhammad~Rehan Siddiqi, Prajvi Saxena, Himanshu Gupta, Tony Fruhauff, Glen Sherman, Mátyás Vincze, Siranut Usawasutsakorn, Dylan Ler, Anil Radhakrishnan, Innocent
  Enyekwe, Sk~Md Salauddin, Jiang Muzhen, Aleksandr Maksapetyan, Vivien Rossbach, Chris Harjadi, Mohsen Bahaloohoreh, Claire Sparrow, Jasdeep Sidhu, Sam Ali, Song Bian, John Lai, Eric Singer, Justine~Leon Uro, Greg Bateman, Mohamed Sayed, Ahmed Menshawy, Darling Duclosel, Dario Bezzi, Yashaswini Jain, Ashley Aaron, Murat Tiryakioglu, Sheeshram Siddh, Keith Krenek, Imad~Ali Shah, Jun Jin, Scott Creighton, Denis Peskoff, Zienab EL-Wasif, Ragavendran~P. V, Michael Richmond, Joseph McGowan, Tejal Patwardhan, Hao-Yu Sun, Ting Sun, Nikola Zubić, Samuele Sala, Stephen Ebert, Jean Kaddour, Manuel Schottdorf, Dianzhuo Wang, Gerol Petruzella, Alex Meiburg, Tilen Medved, Ali ElSheikh, S.~Ashwin Hebbar, Lorenzo Vaquero, Xianjun Yang, Jason Poulos, Vilém Zouhar, Sergey Bogdanik, Mingfang Zhang, Jorge Sanz-Ros, David Anugraha, Yinwei Dai, Anh~N. Nhu, Xue Wang, Ali~Anil Demircali, Zhibai Jia, Yuyin Zhou, Juncheng Wu, Mike He, Nitin Chandok, Aarush Sinha, Gaoxiang Luo, Long Le, Mickaël Noyé, Michał Perełkiewicz,
  Ioannis Pantidis, Tianbo Qi, Soham~Sachin Purohit, Letitia Parcalabescu, Thai-Hoa Nguyen, Genta~Indra Winata, Edoardo~M. Ponti, Hanchen Li, Kaustubh Dhole, Jongee Park, Dario Abbondanza, Yuanli Wang, Anupam Nayak, Diogo~M. Caetano, Antonio A. W.~L. Wong, Maria~del Rio-Chanona, Dániel Kondor, Pieter Francois, Ed~Chalstrey, Jakob Zsambok, Dan Hoyer, Jenny Reddish, Jakob Hauser, Francisco-Javier Rodrigo-Ginés, Suchandra Datta, Maxwell Shepherd, Thom Kamphuis, Qizheng Zhang, Hyunjun Kim, Ruiji Sun, Jianzhu Yao, Franck Dernoncourt, Satyapriya Krishna, Sina Rismanchian, Bonan Pu, Francesco Pinto, Yingheng Wang, Kumar Shridhar, Kalon~J. Overholt, Glib Briia, Hieu Nguyen, David, Soler Bartomeu, Tony~CY Pang, Adam Wecker, Yifan Xiong, Fanfei Li, Lukas~S. Huber, Joshua Jaeger, Romano~De Maddalena, Xing~Han Lù, Yuhui Zhang, Claas Beger, Patrick Tser~Jern Kon, Sean Li, Vivek Sanker, Ming Yin, Yihao Liang, Xinlu Zhang, Ankit Agrawal, Li~S. Yifei, Zechen Zhang, Mu~Cai, Yasin Sonmez, Costin Cozianu, Changhao Li, Alex
  Slen, Shoubin Yu, Hyun~Kyu Park, Gabriele Sarti, Marcin Briański, Alessandro Stolfo, Truong~An Nguyen, Mike Zhang, Yotam Perlitz, Jose Hernandez-Orallo, Runjia Li, Amin Shabani, Felix Juefei-Xu, Shikhar Dhingra, Orr Zohar, My~Chiffon Nguyen, Alexander Pondaven, Abdurrahim Yilmaz, Xuandong Zhao, Chuanyang Jin, Muyan Jiang, Stefan Todoran, Xinyao Han, Jules Kreuer, Brian Rabern, Anna Plassart, Martino Maggetti, Luther Yap, Robert Geirhos, Jonathon Kean, Dingsu Wang, Sina Mollaei, Chenkai Sun, Yifan Yin, Shiqi Wang, Rui Li, Yaowen Chang, Anjiang Wei, Alice Bizeul, Xiaohan Wang, Alexandre~Oliveira Arrais, Kushin Mukherjee, Jorge Chamorro-Padial, Jiachen Liu, Xingyu Qu, Junyi Guan, Adam Bouyamourn, Shuyu Wu, Martyna Plomecka, Junda Chen, Mengze Tang, Jiaqi Deng, Shreyas Subramanian, Haocheng Xi, Haoxuan Chen, Weizhi Zhang, Yinuo Ren, Haoqin Tu, Sejong Kim, Yushun Chen, Sara~Vera Marjanović, Junwoo Ha, Grzegorz Luczyna, Jeff~J. Ma, Zewen Shen, Dawn Song, Cedegao~E. Zhang, Zhun Wang, Gaël Gendron, Yunze Xiao,
  Leo Smucker, Erica Weng, Kwok~Hao Lee, Zhe Ye, Stefano Ermon, Ignacio~D. Lopez-Miguel, Theo Knights, Anthony Gitter, Namkyu Park, Boyi Wei, Hongzheng Chen, Kunal Pai, Ahmed Elkhanany, Han Lin, Philipp~D. Siedler, Jichao Fang, Ritwik Mishra, Károly Zsolnai-Fehér, Xilin Jiang, Shadab Khan, Jun Yuan, Rishab~Kumar Jain, Xi~Lin, Mike Peterson, Zhe Wang, Aditya Malusare, Maosen Tang, Isha Gupta, Ivan Fosin, Timothy Kang, Barbara Dworakowska, Kazuki Matsumoto, Guangyao Zheng, Gerben Sewuster, Jorge~Pretel Villanueva, Ivan Rannev, Igor Chernyavsky, Jiale Chen, Deepayan Banik, Ben Racz, Wenchao Dong, Jianxin Wang, Laila Bashmal, Duarte~V. Gonçalves, Wei Hu, Kaushik Bar, Ondrej Bohdal, Atharv~Singh Patlan, Shehzaad Dhuliawala, Caroline Geirhos, Julien Wist, Yuval Kansal, Bingsen Chen, Kutay Tire, Atak~Talay Yücel, Brandon Christof, Veerupaksh Singla, Zijian Song, Sanxing Chen, Jiaxin Ge, Kaustubh Ponkshe, Isaac Park, Tianneng Shi, Martin~Q. Ma, Joshua Mak, Sherwin Lai, Antoine Moulin, Zhuo Cheng, Zhanda Zhu, Ziyi
  Zhang, Vaidehi Patil, Ketan Jha, Qiutong Men, Jiaxuan Wu, Tianchi Zhang, Bruno~Hebling Vieira, Alham~Fikri Aji, Jae-Won Chung, Mohammed Mahfoud, Ha~Thi Hoang, Marc Sperzel, Wei Hao, Kristof Meding, Sihan Xu, Vassilis Kostakos, Davide Manini, Yueying Liu, Christopher Toukmaji, Jay Paek, Eunmi Yu, Arif~Engin Demircali, Zhiyi Sun, Ivan Dewerpe, Hongsen Qin, Roman Pflugfelder, James Bailey, Johnathan Morris, Ville Heilala, Sybille Rosset, Zishun Yu, Peter~E. Chen, Woongyeong Yeo, Eeshaan Jain, Ryan Yang, Sreekar Chigurupati, Julia Chernyavsky, Sai~Prajwal Reddy, Subhashini Venugopalan, Hunar Batra, Core~Francisco Park, Hieu Tran, Guilherme Maximiano, Genghan Zhang, Yizhuo Liang, Hu~Shiyu, Rongwu Xu, Rui Pan, Siddharth Suresh, Ziqi Liu, Samaksh Gulati, Songyang Zhang, Peter Turchin, Christopher~W. Bartlett, Christopher~R. Scotese, Phuong~M. Cao, Aakaash Nattanmai, Gordon McKellips, Anish Cheraku, Asim Suhail, Ethan Luo, Marvin Deng, Jason Luo, Ashley Zhang, Kavin Jindel, Jay Paek, Kasper Halevy, Allen Baranov,
  Michael Liu, Advaith Avadhanam, David Zhang, Vincent Cheng, Brad Ma, Evan Fu, Liam Do, Joshua Lass, Hubert Yang, Surya Sunkari, Vishruth Bharath, Violet Ai, James Leung, Rishit Agrawal, Alan Zhou, Kevin Chen, Tejas Kalpathi, Ziqi Xu, Gavin Wang, Tyler Xiao, Erik Maung, Sam Lee, Ryan Yang, Roy Yue, Ben Zhao, Julia Yoon, Sunny Sun, Aryan Singh, Ethan Luo, Clark Peng, Tyler Osbey, Taozhi Wang, Daryl Echeazu, Hubert Yang, Timothy Wu, Spandan Patel, Vidhi Kulkarni, Vijaykaarti Sundarapandiyan, Ashley Zhang, Andrew Le, Zafir Nasim, Srikar Yalam, Ritesh Kasamsetty, Soham Samal, Hubert Yang, David Sun, Nihar Shah, Abhijeet Saha, Alex Zhang, Leon Nguyen, Laasya Nagumalli, Kaixin Wang, Alan Zhou, Aidan Wu, Jason Luo, Anwith Telluri, Summer Yue, Alexandr Wang, and Dan Hendrycks.
\newblock Humanity's {Last} {Exam}, April 2025.
\newblock arXiv:2501.14249 [cs].

\bibitem{quan_codeelo_2025}
Shanghaoran Quan, Jiaxi Yang, Bowen Yu, Bo~Zheng, Dayiheng Liu, An~Yang, Xuancheng Ren, Bofei Gao, Yibo Miao, Yunlong Feng, Zekun Wang, Jian Yang, Zeyu Cui, Yang Fan, Yichang Zhang, Binyuan Hui, and Junyang Lin.
\newblock {CodeElo}: {Benchmarking} {Competition}-level {Code} {Generation} of {LLMs} with {Human}-comparable {Elo} {Ratings}, January 2025.
\newblock arXiv:2501.01257 [cs] version: 2.

\bibitem{que_hellobench_2024}
Haoran Que, Feiyu Duan, Liqun He, Yutao Mou, Wangchunshu Zhou, Jiaheng Liu, Wenge Rong, Zekun~Moore Wang, Jian Yang, Ge~Zhang, Junran Peng, Zhaoxiang Zhang, Songyang Zhang, and Kai Chen.
\newblock {HelloBench}: {Evaluating} {Long} {Text} {Generation} {Capabilities} of {Large} {Language} {Models}, September 2024.
\newblock arXiv:2409.16191 [cs].

\bibitem{schmidgall_agent_2025}
Samuel Schmidgall, Yusheng Su, Ze~Wang, Ximeng Sun, Jialian Wu, Xiaodong Yu, Jiang Liu, Zicheng Liu, and Emad Barsoum.
\newblock Agent {Laboratory}: {Using} {LLM} {Agents} as {Research} {Assistants}, January 2025.

\bibitem{shao_assisting_2024}
Yijia Shao, Yucheng Jiang, Theodore~A. Kanell, Peter Xu, Omar Khattab, and Monica~S. Lam.
\newblock Assisting in {Writing} {Wikipedia}-like {Articles} {From} {Scratch} with {Large} {Language} {Models}, April 2024.
\newblock arXiv:2402.14207 [cs].

\bibitem{wang_mobileagentbench_2024}
Luyuan Wang, Yongyu Deng, Yiwei Zha, Guodong Mao, Qinmin Wang, Tianchen Min, Wei Chen, and Shoufa Chen.
\newblock {MobileAgentBench}: {An} {Efficient} and {User}-{Friendly} {Benchmark} for {Mobile} {LLM} {Agents}, June 2024.
\newblock arXiv:2406.08184 [cs].

\bibitem{wang_survey_2025}
Wenxuan Wang, Zizhan Ma, Zheng Wang, Chenghan Wu, Wenting Chen, Xiang Li, and Yixuan Yuan.
\newblock A {Survey} of {LLM}-based {Agents} in {Medicine}: {How} far are we from {Baymax}?, February 2025.
\newblock arXiv:2502.11211 [cs] version: 1.

\bibitem{wei_browsecomp_2025}
Jason Wei, Zhiqing Sun, Spencer Papay, Scott McKinney, Jeffrey Han, Isa Fulford, Hyung~Won Chung, Alex~Tachard Passos, William Fedus, and Amelia Glaese.
\newblock {BrowseComp}: {A} {Simple} {Yet} {Challenging} {Benchmark} for {Browsing} {Agents}, April 2025.
\newblock arXiv:2504.12516 [cs].

\bibitem{wettig_organize_2025}
Alexander Wettig, Kyle Lo, Sewon Min, Hannaneh Hajishirzi, Danqi Chen, and Luca Soldaini.
\newblock Organize the {Web}: {Constructing} {Domains} {Enhances} {Pre}-{Training} {Data} {Curation}, February 2025.
\newblock arXiv:2502.10341 [cs].

\bibitem{wu_webwalker_2025}
Jialong Wu, Wenbiao Yin, Yong Jiang, Zhenglin Wang, Zekun Xi, Runnan Fang, Linhai Zhang, Yulan He, Deyu Zhou, Pengjun Xie, and Fei Huang.
\newblock {WebWalker}: {Benchmarking} {LLMs} in {Web} {Traversal}, January 2025.
\newblock arXiv:2501.07572 [cs].

\bibitem{wu_writingbench_2025}
Yuning Wu, Jiahao Mei, Ming Yan, Chenliang Li, Shaopeng Lai, Yuran Ren, Zijia Wang, Ji~Zhang, Mengyue Wu, Qin Jin, and Fei Huang.
\newblock {WritingBench}: {A} {Comprehensive} {Benchmark} for {Generative} {Writing}, March 2025.
\newblock arXiv:2503.05244 [cs].

\bibitem{xiao_csr-bench_2025}
Yijia Xiao, Runhui Wang, Luyang Kong, Davor Golac, and Wei Wang.
\newblock {CSR}-{Bench}: {Benchmarking} {LLM} {Agents} in {Deployment} of {Computer} {Science} {Research} {Repositories}.
\newblock In Luis Chiruzzo, Alan Ritter, and Lu~Wang, editors, {\em Proceedings of the 2025 {Conference} of the {Nations} of the {Americas} {Chapter} of the {Association} for {Computational} {Linguistics}: {Human} {Language} {Technologies} ({Volume} 1: {Long} {Papers})}, pages 12705--12723, Albuquerque, New Mexico, April 2025. Association for Computational Linguistics.

\bibitem{yan_berkeley_2024}
Fanjia Yan, Huanzhi Mao, Charlie Cheng-Jie Ji, Tianjun Zhang, Shishir~G. Patil, Ion Stoica, and Joseph~E. Gonzalez.
\newblock Berkeley {Function} {Calling} {Leaderboard}, February 2024.
\newblock bibtex[howpublished] = \{{\textbackslash}url\{https://gorilla.cs.berkeley.edu/blogs/8\_berkeley\_function\_calling\_leaderboard.html\}\} citationkey: berkeley-function-calling-leaderboard.

\bibitem{yang_swe-agent_2024}
John Yang, Carlos~E. Jimenez, Alexander Wettig, Kilian Lieret, Shunyu Yao, Karthik Narasimhan, and Ofir Press.
\newblock {SWE}-agent: {Agent}-{Computer} {Interfaces} {Enable} {Automated} {Software} {Engineering}, November 2024.
\newblock arXiv:2405.15793 [cs].

\bibitem{yao_second_nodate}
Shunyu Yao.
\newblock The {Second} {Half}.

\bibitem{zheng_judging_2023}
Lianmin Zheng, Wei-Lin Chiang, Ying Sheng, Siyuan Zhuang, Zhanghao Wu, Yonghao Zhuang, Zi~Lin, Zhuohan Li, Dacheng Li, Eric~P. Xing, Hao Zhang, Joseph~E. Gonzalez, and Ion Stoica.
\newblock Judging {LLM}-as-a-{Judge} with {MT}-{Bench} and {Chatbot} {Arena}, December 2023.
\newblock arXiv:2306.05685 [cs].

\bibitem{zheng_deepresearcher_2025}
Yuxiang Zheng, Dayuan Fu, Xiangkun Hu, Xiaojie Cai, Lyumanshan Ye, Pengrui Lu, and Pengfei Liu.
\newblock {DeepResearcher}: {Scaling} {Deep} {Research} via {Reinforcement} {Learning} in {Real}-world {Environments}, April 2025.
\newblock arXiv:2504.03160 [cs].

\bibitem{zhou_browsecomp-zh_2025}
Peilin Zhou, Bruce Leon, Xiang Ying, Can Zhang, Yifan Shao, Qichen Ye, Dading Chong, Zhiling Jin, Chenxuan Xie, Meng Cao, Yuxin Gu, Sixin Hong, Jing Ren, Jian Chen, Chao Liu, and Yining Hua.
\newblock {BrowseComp}-{ZH}: {Benchmarking} {Web} {Browsing} {Ability} of {Large} {Language} {Models} in {Chinese}, May 2025.
\newblock arXiv:2504.19314 [cs].

\bibitem{zhou_webarena_2024}
Shuyan Zhou, Frank~F. Xu, Hao Zhu, Xuhui Zhou, Robert Lo, Abishek Sridhar, Xianyi Cheng, Tianyue Ou, Yonatan Bisk, Daniel Fried, Uri Alon, and Graham Neubig.
\newblock {WebArena}: {A} {Realistic} {Web} {Environment} for {Building} {Autonomous} {Agents}, April 2024.
\newblock arXiv:2307.13854 [cs].

\bibitem{zhuo_bigcodebench_2025}
Terry~Yue Zhuo, Minh~Chien Vu, Jenny Chim, Han Hu, Wenhao Yu, Ratnadira Widyasari, Imam Nur~Bani Yusuf, Haolan Zhan, Junda He, Indraneil Paul, Simon Brunner, Chen Gong, Thong Hoang, Armel~Randy Zebaze, Xiaoheng Hong, Wen-Ding Li, Jean Kaddour, Ming Xu, Zhihan Zhang, Prateek Yadav, Naman Jain, Alex Gu, Zhoujun Cheng, Jiawei Liu, Qian Liu, Zijian Wang, Binyuan Hui, Niklas Muennighoff, David Lo, Daniel Fried, Xiaoning Du, Harm~de Vries, and Leandro~Von Werra.
\newblock {BigCodeBench}: {Benchmarking} {Code} {Generation} with {Diverse} {Function} {Calls} and {Complex} {Instructions}, April 2025.
\newblock arXiv:2406.15877 [cs].

\end{thebibliography}

%%%%%%%%%%%%%%%%%%%%%%%%%%%%%%%%%%%%%%%%%%%%%%%%%%%%%%%%%%%%

\newpage
\appendix
\label{sec:appendix}

\section{Limitations}\label{sec:limitations}

While \textbf{DeepResearch Bench} and the RACE \& FACT evaluation framework offer the first comprehensive assessment of Deep Research Agents, several limitations remain.

\paragraph{Benchmark scale.} Creating tasks that authentically reflect real research challenges requires substantial expertise and effort. Each of our 100 tasks was developed by a verified PhD-level expert who underwent orientation on our task specifications and participated in multiple rounds of refinement. Following rigorous internal quality screening, only superior tasks were included. This meticulous process ensures high quality but inevitably constrains the dataset size. Expanding the benchmark while maintaining our quality standards would enhance statistical robustness and topic coverage, representing a key priority for future development.

\paragraph{Domain coverage bias.} Despite our team's breadth of expertise, inadvertent biases may have been introduced during the curation and selection process. We implemented multiple review layers and explicit acceptance criteria to mitigate this risk, yet we cannot guarantee complete impartiality. Future iterations will incorporate additional external reviewers with specialized domain knowledge to achieve a more balanced representation across fields.

\paragraph{Human evaluation throughput.} The assessment of research reports presents inherent challenges: a single report may contain dozens of pages with dense technical content and sophisticated analysis. Even for PhD-level evaluators, a thorough review typically requires 30–60 minutes, with the evaluation of multiple reports multiplying this burden. For the current study, we limited human evaluation to three experts per task, covering 50 tasks across four agents (totaling 600 reports). While sufficient to validate the RACE framework, this sample remains modest and susceptible to individual variability. We intend to conduct more extensive annotation campaigns to strengthen statistical confidence and refine our automated metrics, thereby maintaining alignment with human judgment.

\paragraph{Broader Impact} % Or use \section* or just as a paragraph

Our work on DeepResearch Bench aims to significantly accelerate research and innovation by providing a robust framework for developing more capable Deep Research Agents (DRAs). This can democratize access to advanced research methodologies, empowering a wider range of users across various domains. However, the advancement of powerful DRAs also presents societal challenges that warrant careful consideration. These include the potential for generating sophisticated misinformation if not rigorously validated, the risk of over-reliance, which could lead to a decline in critical thinking and manual research skills, and the possibility of amplifying biases inherent in the underlying models and the vast web data they process. DeepResearch Bench, with its multifaceted evaluation that includes factual grounding (FACT) and quality assessment (RACE), is designed as a step towards mitigating such risks by fostering the development of more reliable, transparent, and responsible AI agents. Continuous vigilance, ongoing research into AI safety and ethics, and multi-stakeholder engagement will be crucial to ensure these powerful tools are developed and deployed in a manner that benefits society while minimizing potential harms.

\section{Definitions of Evaluation Dimensions in RACE}
\label{app:dimension_definitions}

The RACE framework evaluates research reports based on four top-level dimensions. Their definitions are provided in Table~\ref{tab:dimension_definitions}.

\begin{table}[htbp]
\centering
\caption{Definitions of Core Evaluation Dimensions for Report Quality}
\label{tab:dimension_definitions}
\small                       % ↓ 关键：宽度用 \linewidth 而不是 \textwidth
\begin{tabularx}{\linewidth}{@{}lX@{}}  
\toprule
\textbf{Dimension} & \textbf{Description} \\
\midrule
\textbf{Comprehensiveness} (\textsc{Comp}) &
Article covers key areas of the industry, ensures overall understanding, and does not omit important parts. \\[2pt]

\textbf{Insight/Depth} (\textsc{Depth}) &
Article deeply analyzes causes, impacts, and trends, providing valuable insights. \\[2pt]

\textbf{Instruction\hyp{}Following/Relevance} (\textsc{Inst}) &
Article closely follows the research topic and directly answers questions. \\[2pt]

\textbf{Readability} (\textsc{Read}) &
Article has a clear structure, fluent language, and is easy to understand. \\
\bottomrule
\end{tabularx}
\end{table}

\section{Judge LLM Selection for the FACT Framework}
\label{app:fact_judge_llm_selection}

The FACT framework employs a Judge LLM for crucial automated steps: Statement-URL Pair Extraction and Deduplication, followed by Support Judgment. The selection of Judge LLM is pivotal, aiming to balance evaluation accuracy with operational costs, especially given the significant token consumption inherent in these processes. To determine an optimal model for these tasks, we specifically evaluated \texttt{Gemini-2.5-Flash}. Its judgments were compared against human evaluations on a randomly sampled set of 100 statement-URL pairs derived from our benchmark tasks. This comparison demonstrated strong agreement with human annotators: \texttt{Gemini-2.5-Flash}'s judgment aligned with human 'support' determinations in 96\% of cases and with 'not support' determinations in 92\% of cases.

We further find out that the accuracy of \texttt{Gemini-2.5-Flash} in these FACT-specific evaluation steps is very close to that of \texttt{Gemini-2.5-Pro}. The operations within the FACT framework (such as extracting statements from full reports and analyzing webpage content for support) are known to be token-intensive, making cost-effectiveness a critical consideration. Since \texttt{Gemini-2.5-Flash} demonstrated comparable accuracy to \texttt{Gemini-2.5-Pro} for these specific tasks but at a more advantageous cost, we select \texttt{Gemini-2.5-Flash} as the Judge LLM for the FACT framework. This choice enables us to maintain high evaluation reliability while managing operational costs effectively.

\section{Data Collection Timeframes for Evaluated Models}
\label{app:data_collection_timeframes}

The data for the commercial models evaluated in this paper were collected during specific timeframes in 2025, as detailed in Table~\ref{tab:data_collection_dates}. These dates indicate when the model outputs used in our experiments were generated.

% Ensure you have \usepackage{booktabs} in your document preamble
\begin{table}[htbp]
\centering
\caption{Data Collection Timeframes for Evaluated Models (2025)}
\label{tab:data_collection_dates}
\begin{tabular}{@{}ll@{}} % Two left-aligned columns
\toprule
\textbf{Model Category / Provider Group} & \textbf{Data Collection Date Range} \\
\midrule
\multicolumn{2}{l}{\textit{Deep Research Agents (DRAs)}} \\ % Subheading for DRAs
OpenAI Deep Research            & April 1 -- May 8 \\
Gemini 2.5 Pro Deep Research    & April 27 -- April 29 \\
Perplexity Deep Research        & April 1 -- April 29 \\
Grok Deeper Search              & April 27 -- April 29 \\
\addlinespace % Adds a little vertical space (from booktabs package)
\multicolumn{2}{l}{\textit{LLM with Search Tools (Grouped by Provider)}} \\ % Subheading for other models
Claude Models (w/Search)        & May 12 -- May 13 \\
Perplexity Models    & May 11 -- May 12 \\
GPT Models (w/Search)           & May 11 -- May 12 \\
Gemini Models (Grounding/w/Search) & May 12 -- May 13 \\
\bottomrule
\end{tabular}
\end{table}

\section{Detailed Calculation of Citation Metrics}
\label{app:citation_metrics_calculation}

This appendix provides the detailed definitions and calculation methods for the Citation Accuracy (C. Acc.) and Average Effective Citations per Task (E. Cit.) metrics used within the FACT framework.

Let $T$ denote the set of all tasks in the benchmark, and $|T|$ be the total number of tasks.
For each task $t \in T$:
\begin{itemize}
    \item Let $U_t$ be the set of unique statement-URL pairs extracted for task $t$ after the deduplication process.
    \item Let $N_{u,t} = |U_t|$ be the total number of unique statement-URL pairs for task $t$ that undergo support judgment.
    \item For each statement-URL pair in $U_t$, a support judgment is rendered, which can be either 'support' or 'not support'.
    \item Let $N_{s,t}$ be the number of statement-URL pairs that are judged as 'support' for task $t$.
\end{itemize}

\subsection{Citation Accuracy (C. Acc.)}
Citation Accuracy (C. Acc.) assesses the Deep Research Agent's (DRA) ability to accurately apply retrieved information for precise statements. It is calculated by first determining the proportion of 'support' statement-URL pairs for each individual task, and then averaging these per-task accuracies across all tasks in the benchmark.

The accuracy for a single task $t$, denoted as $Acc_t$, is defined as:
\begin{equation}
Acc_t = \begin{cases} 
            \frac{N_{s,t}}{N_{u,t}} & \text{if } N_{u,t} > 0 \\
            0 & \text{if } N_{u,t} = 0 
            \end{cases}
\end{equation}
This definition ensures that tasks for which the agent produces no citable statements (i.e., $N_{u,t} = 0$) contribute an accuracy of 0 to the overall average, reflecting a failure to provide supported information for that task.

The overall \textbf{Citation Accuracy (C. Acc.)} is then computed as the average of these per-task accuracies over all tasks in the benchmark:
\begin{equation}
\text{C. Acc.} = \frac{1}{|T|} \sum_{t \in T} Acc_t
\end{equation}

\subsection{Average Effective Citations per Task (E. Cit.)}
Average Effective Citations per Task (E. Cit.) evaluates, on average, how much useful and relevant information the agent retrieves and correctly supports with evidence for each task. It is computed by summing the total number of 'support' statement-URL pairs across all tasks and then dividing by the total number of tasks in the benchmark.

The \textbf{Average Effective Citations per Task (E. Cit.)} is calculated as:
\begin{equation}
\text{E. Cit.} = \frac{\sum_{t \in T} N_{s,t}}{|T|}
\end{equation}
This metric provides a direct measure of the average quantity of verifiably supported statements an agent generates per task.

\section{Detailed Calculation of Human Consistency Metrics}
\label{app:human_consistency_metrics_detailed}

This appendix provides the detailed calculation methods for the four metrics used to validate the consistency between our RACE framework and human judgment, as introduced in Section~\ref{subsubsec:metrics}.

\subsection{Pairwise Agreement Rate}
The Pairwise Agreement Rate measures the proportion of report pairs (across all tasks) where the evaluation method's preference matches the human preference.

For each of the $N_t=50$ tasks in our study, four generated deep research reports result in $N_p=\binom{4}{2}=6$ unique pairs per task. Human preference for each pair (e.g., Report A is better than Report B, or they are tied) is established from the average overall scores assigned to each report by three domain experts.

Let $\text{I}(t, p)$ be an indicator function for task $t$ and pair $p$:
\begin{equation}
\text{I}(t, p) = 
\begin{cases} 
1 & \text{if the method's preference matches the human preference for pair } p \text{ of task } t \\
0 & \text{otherwise.}
\end{cases}
\end{equation}
The Pairwise Agreement Rate is then calculated as:
\begin{equation}
\text{Pairwise Agreement Rate} = \frac{\sum_{t=1}^{N_t} \sum_{p=1}^{N_p} \text{I}(t, p)}{N_t \times N_p}.
\end{equation}
This metric reflects the evaluation method's reliability in replicating human comparative judgments.

\subsection{Overall Pearson Correlation}
This metric quantifies the linear relationship between average model scores from the evaluation method and those from human experts, aggregated across all $N_t=50$ tasks.

Let $X$ be a vector of average scores per model (e.g., for the different DRAs evaluated) obtained from our method, aggregated across all tasks.
Let $Y$ be the corresponding vector of average scores per model, obtained from human experts and aggregated across all tasks.
The Overall Pearson Correlation is the standard Pearson correlation coefficient $r(X,Y)$ calculated between these two vectors. This reflects the overall score correlation between the method and human experts for the evaluated DRAs.

\subsection{Filtered Average Pearson Correlation}
This metric calculates the average of per-task Pearson correlations ($r_t$) between the method's scores and mean human scores, specifically on tasks where human judgment is more consistent.

Given that human scores are from a limited number of experts ($k=3$) for the $n=4$ reports per task, expert inconsistencies can affect task-level metric stability. To mitigate this, tasks are filtered based on inter-rater reliability using the Intraclass Correlation Coefficient (ICC). For each task, ICC(1,1) (a one-way random effects model) is computed from the $k=3$ human experts' scores for the $n=4$ reports:
\begin{equation}
\text{ICC}(1,1) = \frac{\text{MSB}-\text{MSW}}{\text{MSB}+(k-1)\text{MSW}},
\end{equation}
where MSB is the mean square between reports and MSW is the mean square within reports. Tasks with poor inter-rater reliability (e.g., $\text{ICC}(1,1) < 0$) are excluded. This yields a filtered subset of $N_{\text{filtered}}$ tasks, denoted as $\mathcal{T}_{\text{filtered}}$ (37 in our experiments).

The Filtered Average Pearson correlation is then the average of per-task Pearson correlations ($r_t$) between the method's scores and mean human scores over $\mathcal{T}_{\text{filtered}}$:
\begin{equation}
\text{Filtered Avg Pearson} = \frac{1}{N_{\text{filtered}}}\sum_{t\in\mathcal{T}_{\text{filtered}}} r_t.
\end{equation}
This procedure provides a more robust assessment of absolute-score correlation.

\subsection{Filtered Average Spearman Correlation}
Using the same filtering method and the subset $\mathcal{T}_{\text{filtered}}$, this metric evaluates model ranking consistency.

For each task $t \in \mathcal{T}_{\text{filtered}}$, the Spearman rank correlation coefficient $\rho_t$ is calculated between model rankings derived from our evaluation method and those from average human scores. The Filtered Average Spearman Correlation is then the average of these $\rho_t$ values:
\begin{equation}
\text{Filtered Avg Spearman} = \frac{1}{N_{\text{filtered}}}\sum_{t\in\mathcal{T}_{\text{filtered}}} \rho_t.
\end{equation}
This reflects how well the method preserves relative model ordering compared to humans, specifically on tasks with more consistent human judgments.

\section{Experiment Detail}
\label{sec: exp detail}

\subsection{Human Evaluation Effort}
\label{subsec: human details}
The human evaluation process required a considerable investment of time to ensure a thorough assessment. On average, each expert annotator spent approximately 1.5 hours per query to evaluate the reports from the four agents. This meticulous process culminated in a total of 225 person-hours of human evaluation across all tasks and annotators. This substantial effort yielded a robust and reliable dataset of human judgments, serving as the basis for our analysis of human consistency.

\subsection{Configuration of LLMs with Web Search Tools}
\label{subsec:llm_search_config} % You can change this label

To ensure a standardized and comparable evaluation environment for Large Language Models (LLMs) equipped with built-in web search tools, the following configurations were uniformly applied:

\begin{itemize}
    \item \textbf{Thinking Budget/Computational Resources:} For models that support a configurable ``thinking budget'' or a similar computational resource limit for generation, this was uniformly set to a high value, equivalent to 16,000 tokens where applicable. This allowed models ample processing capacity for complex queries.

    \item \textbf{Search Context Size:} In cases where models offered a parameter to control the amount of information retrieved and utilized from web searches (e.g., the \texttt{search\_context\_size} option as found in the Perplexity AI API), this was consistently set to ``high''. This configuration aimed to maximize the contextual information available to the LLM from its search activities.

    \item \textbf{Maximum Search Iterations:} The maximum number of web search queries, or ``search turns,'' permitted during the generation process was standardized to five for all LLMs that provided such a configurable limit. This ensured a comparable depth of web exploration across these models.

    \item \textbf{Output Length:} To accommodate potentially comprehensive responses while maintaining consistency, the maximum output token limit for all LLMs was set to 36,000 tokens. If a model's inherent maximum output capacity was less than 36,000 tokens, its specific native maximum limit was adhered to.

    \item \textbf{Citation Formatting and Standardization:} A critical aspect of our methodology was the standardization of citation presentation to facilitate consistent downstream evaluation, particularly when employing frameworks like FACT for factual assessment. Citations as provided by each LLM were parsed in accordance with their respective official API documentation. Subsequently, the generated reports were systematically restructured: citation markers were inserted in the format `[1][2]` at the end of the relevant sentences, and a consolidated 'References' list, compiling all unique cited sources, was appended to the conclusion of each article. This uniform approach to citation structure was essential for equitable and rigorous factual verification.
\end{itemize}

These standardized settings were implemented to minimize variability arising from differing default configurations and to enable a more direct comparison of the models' capabilities in the context of deep research tasks.

\section{Prompt Templates}

\begin{tcolorbox}[
  enhanced,
  breakable,
  colback=ysshallowred,
  colframe=ysdarkred,
  title=Clean Article Prompt,
  fonttitle=\bfseries
]

<system\_role>

You are a professional article editor who is good at cleaning and refining article content.

</system\_role>

<user\_prompt>

Please help me clean the following research article, removing all citation links, citation marks (such as [1], [2], 1, 2, etc. or other complex citation formats), reference lists, footnotes, and ensuring the content is coherent and smooth.
Keep all other original content of the article, removing only the citations. If the content of the citation mark is used as part of a sentence in the article, keep the text content and remove other marks.

Article content:
"\{article\}"

Please return the cleaned article in full, without adding any additional comments or explanations.

</user\_prompt>
\end{tcolorbox}

\vspace{1em}
%========================================
% 2) generate_eval_dimension_weight_prompt
%========================================

\begin{tcolorbox}[
  % enhanced,
  breakable,
  colback=ysshallowpurple,
  colframe=ysdarkpurple,
  title={Generate Dynamic Dimension Weight Prompt},
  fonttitle=\bfseries
]
<system\_role>

You are an experienced research article evaluation expert. You excel at deeply understanding the objectives, challenges, and core value points of specific research tasks, and based on this, setting \textbf{dynamic, reasonable, and well-supported} dimension weights for subsequent article quality assessment.

</system\_role>

<user\_prompt>

There is a deep research task as follows:

<task>

"\{task\_prompt\}"

</task>

<instruction>

\textbf{Background}: The research team will conduct in-depth and comprehensive research based on the `<task>` above and ultimately produce a high-quality research article.

\textbf{Your Task}: As an evaluation expert, you need to set the evaluation criteria weights for this specific `<task>` for our assessment team. The evaluation will be conducted across the following four dimensions:
\begin{enumerate}
    \item \textbf{Comprehensiveness:} The breadth, depth, and relevance of information coverage.
    \item \textbf{Insight:} The depth, originality, logic, and value of the analysis and conclusions.
    \item \textbf{Instruction Following:} Whether the report accurately and completely responds to all requirements and constraints of the task.
    \item \textbf{Readability:} Clarity of structure, fluency of language, effectiveness of data presentation, and overall ease of understanding.
\end{enumerate}

\textbf{Evaluation Formula}: Total Score = Comprehensiveness * Comprehensiveness Weight + Insight * Insight Weight + Instruction Following * Instruction Following Weight + Readability * Readability Weight. (\textbf{Note: The sum of all weights must be exactly 1.0})

\textbf{Core Requirements}:
\begin{enumerate}
    \item \textbf{In-depth Task Analysis}: Carefully study the specific content of the `<task>`, its implicit goals, potential difficulties, and the core value of its outcomes.
    \item \textbf{Dynamic Weight Allocation}: Based on your analysis, assign weights to the four dimensions (use decimals between 0 and 1, e.g., 0.3). \textbf{The key is to understand that different tasks have different focuses, and weights must be flexibly adjusted according to task characteristics, not fixed.}
    \item ...
\end{enumerate}

</instruction>

<examples\_rationale>

The following two examples are provided to demonstrate \textbf{how to adjust evaluation dimension weights and explain the reasons based on changes in task nature}. Please focus on learning the \textbf{thinking logic and analytical methods} in these examples, rather than simply imitating their content or weight values.

</examples\_rationale>

<example\_1>

<task>

"Analyze the feasibility of investing in electric vehicle (EV) charging infrastructure in suburban areas."

</task>

<output>

<analysis>

This task's core is to provide a clear feasibility analysis for a specific investment. The value lies in the thoroughness of the assessment and the practicality of its conclusions. Therefore, evaluation emphasizes insight and comprehensiveness.
\begin{itemize}
    \item \textbf{Insight (0.35):} The task requires a deep analysis of feasibility, including market demand, costs, competition, and regulatory landscape. The quality of the strategic recommendations derived from this analysis is key.
    \item ...
\end{itemize}

</analysis>

<json\_output>

\{\{
    "comprehensiveness": 0.30,
    "insight": 0.35,
    "instruction\_following": 0.20,
    "readability": 0.15
\}\}

</json\_output>

</output>

</example\_1>

Please strictly follow the above instructions and methods. Now, begin your work on the following specific task:

<task>

"\{task\_prompt\}"

</task>

Please output your `<analysis>` and `<json\_output>`.

</user\_prompt>
\end{tcolorbox}

\newpage

%========================================
% 3) generate_eval_criteria_prompt_comp
%========================================
\begin{tcolorbox}[
  enhanced,
  breakable,
  colback=ysshallowpurple,
  colframe=ysdarkpurple,
  title={Generate Comprehensiveness Criteria Prompt},
  fonttitle=\bfseries
]
<system\_role>

You are an experienced research article evaluation expert. You excel at breaking down abstract evaluation dimensions (like "Comprehensiveness") into actionable, clear, and task-specific criteria, assigning appropriate weights and justifications for each.

</system\_role>

<user\_prompt>

\textbf{Background}: We are evaluating a deep research article written for the following task across four dimensions: Comprehensiveness, Insight, Instruction Following, and Readability.
\begin{enumerate}
    \item \textbf{Comprehensiveness:} The breadth, depth, and relevance of information coverage.
    \item \textbf{Insight:} The depth, originality, logic, and value of the analysis and conclusions.
    \item \textbf{Instruction Following:} Whether the report accurately and completely responds to all requirements and constraints of the task.
    \item \textbf{Readability:} Clarity of structure, fluency of language, effectiveness of data presentation, and overall ease of understanding.
\end{enumerate}

<task>

"\{task\_prompt\}"

</task>

<instruction>

\textbf{Your Goal}: For the \textbf{Comprehensiveness} dimension of this research article, develop a set of detailed, specific, and highly task-relevant evaluation criteria. You need to:
\begin{enumerate}
    \item \textbf{Analyze Task}: Deeply analyze the `<task>` to identify key information areas, perspectives, and depths that must be covered to achieve "comprehensiveness."
    \item \textbf{Formulate Criteria}: Based on the analysis, propose specific evaluation criteria items.
    \item \textbf{Explain Rationale}: Provide a brief explanation (`explanation`) for each criterion, stating why it is important for assessing the comprehensiveness of this `<task>`.
    \item ...
\end{enumerate}

\textbf{Core Requirements}:
\begin{enumerate}
    \item \textbf{Task-Centric}: Analysis, criteria, explanations, and weights must directly relate to the core requirements and characteristics of the `<task>`.
    \item \textbf{Well-Justified}: The `<analysis>` section must clearly articulate the overall thinking behind setting these criteria and weights, linking it to the `<task>`. The `explanation` for each criterion must justify its specific relevance.
    \item ...
\end{enumerate}

</instruction>

<example\_rational>

The following example demonstrates \textbf{how to formulate comprehensiveness criteria based on task requirements}. Focus on learning the \textbf{thinking logic and analytical methods} from this example, not just imitating its content or weight values.

</example\_rational>

<example>

<task>

"Analyze the impact of remote work trends on commercial real estate in major US cities and recommend investment strategies."

</task>

<output>

<analysis>

To comprehensively evaluate a research article on "the impact of remote work on commercial real estate in major US cities and recommended investment strategies," considerations must span multiple dimensions. This task has a dual objective: first, to analyze the market impact, and second, to propose actionable investment strategies based on this analysis. Therefore, a comprehensive assessment must ensure the article covers key drivers of change in commercial real estate due to remote work, and thoroughly examines various investment approaches.

Specifically, evaluation criteria need to cover:
\begin{enumerate}
    \item \textbf{Remote Work Trends \& Adoption Data}: Coverage of current and projected remote/hybrid work models, adoption rates across industries and demographics.
    \item \textbf{Impact on Commercial Real Estate Sectors}: Analysis of effects on office, retail, and industrial spaces, including vacancy rates, leasing trends, and property valuations in major US cities.
    \item \textbf{Geographical Variations}: Examination of how impacts differ across various major US cities (e.g., tech hubs vs. financial centers, downtown vs. suburban).
    \item ...
\end{enumerate}

Weight allocation should be balanced between the impact analysis (remote work trends, sector impacts, geographical variations) and the investment strategy section, as both are critical to fulfilling the task. Within impact analysis, specific sector impacts and geographical variations are key to actionable insights.

</analysis>

<json\_output>

[

  \{\{
  
    "criterion": "Analysis of Remote Work Trends and Adoption",
    
    "explanation": "Assesses if the article thoroughly examines current and projected remote/hybrid work models, adoption statistics across different industries, and demographic factors influencing these trends. This forms the basis of the impact analysis.",
    
    "weight": 0.15
    
  \}\},
  
  \{\{
  
    "criterion": "Comprehensive Coverage of CRE Sector Impacts",
    
    "explanation": "Evaluates if the article details the impact on various commercial real estate (CRE) sectors (office, retail, industrial, etc.), including data on vacancy rates, rental trends, and property valuations in major US cities.",
    
    "weight": 0.20
    
  \}\},
  
  \{\{
  
    "criterion": "Examination of Geographical Variations and Nuances",
    
    "explanation": "Checks if the article analyzes how the impact of remote work on CRE differs across various major US cities, considering their unique economic structures and demographic profiles.",
    
    "weight": 0.15
    
  \}\},
  
  \{\{
  
    "criterion": "Discussion of Broader Economic and Social Consequences",
    
    "explanation": "Assesses coverage of wider implications, such as effects on urban planning, transportation, local economies dependent on office workers, and ancillary businesses.",
    
    "weight": 0.10
    
  \}\},
  
  ...
  
]

</json\_output>

</output>

</example>

Please strictly follow the above instructions and methods. Now, begin your work on the following specific task:

<task>

"\{task\_prompt\}"

</task>

Please output your `<analysis>` and `<json\_output>`.

</user\_prompt>
\end{tcolorbox}

\newpage

\begin{tcolorbox}[
  enhanced,
  breakable,
  colback=ysshallowblue,
  colframe=ysdarkblue,
  title={Score Prompt In RACE(Full)},
  fonttitle=\bfseries
]
<system\_role>

You are a strict, meticulous, and objective research article evaluation expert. You excel at using specific assessment criteria to deeply compare two articles on the same task, providing precise scores and clear justifications.

</system\_role>

<user\_prompt>

\textbf{Task Background}

There is a deep research task, and you need to evaluate two research articles written for this task. We will assess the articles across four dimensions: Comprehensiveness, Insight, Instruction Following, and Readability. The content is as follows:

<task>

"\{task\_prompt\}"

</task>

\textbf{Articles to Evaluate}

<article\_1>

"\{article\_1\}"

</article\_1>

<article\_2>

"\{article\_2\}"

</article\_2>

\textbf{Evaluation Criteria}
Now, you need to evaluate and compare these two articles based on the following \textbf{evaluation criteria list}, providing comparative analysis and scoring each on a scale of 0-10. Each criterion includes an explanation, please understand carefully.

<criteria\_list>

\{criteria\_list\}

</criteria\_list>

<Instruction>

\textbf{Your Task}

Please strictly evaluate and compare `<article\_1>` and `<article\_2>` based on \textbf{each criterion} in the `<criteria\_list>`. You need to:

\begin{enumerate}
    \item \textbf{Analyze Each Criterion}: Consider how each article fulfills the requirements of each criterion.
    \item \textbf{Comparative Evaluation}: Analyze how the two articles perform on each criterion, referencing the content and criterion explanation.
    \item \textbf{Score Separately}: Based on your comparative analysis, score each article on each criterion (0-10 points).
\end{enumerate}

\textbf{Scoring Rules}

For each criterion, score both articles on a scale of 0-10 (continuous values). The score should reflect the quality of performance on that criterion:

\begin{itemize}
    \item 0-2 points: Very poor performance. Almost completely fails to meet the criterion requirements.
    \item 2-4 points: Poor performance. Minimally meets the criterion requirements with significant deficiencies.
    \item 4-6 points: Average performance. Basically meets the criterion requirements, neither good nor bad.
    \item 6-8 points: Good performance. Largely meets the criterion requirements with notable strengths.
    \item 8-10 points: Excellent/outstanding performance. Fully meets or exceeds the criterion requirements.
\end{itemize}

\textbf{Output Format Requirements}

Please \textbf{strictly} follow the `<output\_format>` below for each criterion evaluation. \textbf{Do not include any other unrelated content, introduction, or summary}. Start with "Standard 1" and proceed sequentially through all criteria:

</Instruction>

<output\_format>

\{\{

    "comprehensiveness": [
    
        \{\{
        
            "criterion": [Text content of the first comprehensiveness evaluation criterion],
            
            "analysis": [Comparative analysis],
            
            "article\_1\_score": [Continuous score 0-10],
            
            "article\_2\_score": [Continuous score 0-10]
            
        \}\},
        
        \{\{
        
            "criterion": [Text content of the second comprehensiveness 
            evaluation criterion],
            
            "analysis": [Comparative analysis],
            
            "article\_1\_score": [Continuous score 0-10],
            
            "article\_2\_score": [Continuous score 0-10]
            
        \}\},
        
        ...
        
    ],
    
    "insight": [
    
        \{\{
        
            "criterion": [Text content of the first insight evaluation criterion],
            
            "analysis": [Comparative analysis],
            
            "article\_1\_score": [Continuous score 0-10],
            
            "article\_2\_score": [Continuous score 0-10]
            
        \}\},
        
        ...
        
    ],
    
    ...
    
\}\}

</output\_format>

Now, please evaluate the two articles based on the research task and criteria, providing detailed comparative analysis and scores according to the requirements above. Ensure your output follows the specified `<output\_format>` and that the JSON format is parsable, with all characters that might cause JSON parsing errors properly escaped.

</user\_prompt>
\end{tcolorbox}

\newpage

\begin{tcolorbox}[
  enhanced,
  breakable,
  colback=ysshallowblue,
  colframe=ysdarkblue,
  title={Static Score Prompt},
  fonttitle=\bfseries
]
<system\_role>

You are a strict, meticulous, and objective research article evaluation expert.

You excel at using specific assessment criteria to deeply compare two articles on the same task, providing precise scores and clear justifications.

</system\_role>

<user\_prompt>

\textbf{Task Background}

There is a deep research task, and you need to evaluate two research articles written for this task.

We will assess the articles across four dimensions: Comprehensiveness, Insight, Instruction Following, and Readability.

The content is as follows:

<task>

"\{task\_prompt\}"

</task>

\textbf{Articles to Evaluate}

<article\_1>

"\{article\_1\}"

</article\_1>

<article\_2>

"\{article\_2\}"

</article\_2>

\textbf{Evaluation Criteria}

Now, you need to evaluate and compare these two articles based on the following \textbf{fixed evaluation criteria list}, providing comparative analysis and scoring each on a scale of 0-10.

Each criterion includes an explanation, please understand carefully.

<criteria\_list>

\textbf{Comprehensiveness}

[
  \{\{
    "criterion": "Information Coverage Breadth",
    
    "explanation": "Evaluates whether the article covers all key areas and aspects related to the topic without omitting important information.",
    
    "weight": 0.25
    
  \}\},
  
  \{\{
  
    "criterion": "Information Depth and Detail",
    
    "explanation": "Evaluates whether the article provides sufficiently detailed information rather than just surface-level overviews.",
    
    "weight": 0.25
    
  \}\},

  ...
]

\textbf{Insight}

[
  \{\{
    "criterion": "Analysis Depth and Originality",
    
    "explanation": "Evaluates whether the article provides deep analysis and original insights rather than simply repeating known information.",
    
    "weight": 0.25
    
  \}\},
  
  ...
  
]

...

</criteria\_list>

<Instruction>

\textbf{Your Task}

Please strictly evaluate and compare `<article\_1>` and `<article\_2>` based on \textbf{each criterion} in the `<criteria\_list>`.

You need to:
\begin{enumerate}
    \item \textbf{Analyze Each Criterion}: Consider how each article fulfills the requirements of each criterion.
    \item \textbf{Comparative Evaluation}: Analyze how the two articles perform on each criterion, referencing the content and criterion explanation.
    \item \textbf{Score Separately}: Based on your comparative analysis, score each article on each criterion (0-10 points).
\end{enumerate}

\textbf{Scoring Rules}

For each criterion, score both articles on a scale of 0-10 (continuous values).

The score should reflect the quality of performance on that criterion:
\begin{itemize}
    \item 0-2 points: Very poor performance. Almost completely fails to meet the criterion requirements.
    \item 2-4 points: Poor performance. Minimally meets the criterion requirements with significant deficiencies.
    \item 4-6 points: Average performance. Basically meets the criterion requirements, neither good nor bad.
    \item 6-8 points: Good performance. Largely meets the criterion requirements with notable strengths.
    \item 8-10 points: Excellent/outstanding performance. Fully meets or exceeds the criterion requirements.
\end{itemize}

\textbf{Output Format Requirements}

Please \textbf{strictly} follow the `<output\_format>` below for each criterion evaluation.

\textbf{Do not include any other unrelated content, introduction, or summary}.

Start with "Standard 1" and proceed sequentially through all criteria:

</Instruction>

<output\_format>

\{\{

    "comprehensiveness": [
    
        \{\{
        
            "criterion": [Text content of the first comprehensiveness evaluation criterion],
            
            "analysis": [Comparative analysis],
            
            "article\_1\_score": [Continuous score 0-10],
            
            "article\_2\_score": [Continuous score 0-10]
            
        \}\},
        
        \{\{
        
            "criterion": [Text content of the second comprehensiveness evaluation criterion],
            
            "analysis": [Comparative analysis],
            
            "article\_1\_score": [Continuous score 0-10],
            
            "article\_2\_score": [Continuous score 0-10]
            
        \}\},
        
        ...
        
    ],
    "insight": [
    
        \{\{
        
            "criterion": [Text content of the first insight evaluation criterion],
            
            "analysis": [Comparative analysis],
            
            "article\_1\_score": [Continuous score 0-10],
            
            "article\_2\_score": [Continuous score 0-10]
            
        \}\},
        
        ...
        
    ],
    
    ...
    
\}\}

</output\_format>

Now, please evaluate the two articles based on the research task and criteria, providing detailed comparative analysis and scores according to the requirements above.

Ensure your output follows the specified `<output\_format>` and that the JSON format is parsable, with all characters that might cause JSON parsing errors properly escaped.

</user\_prompt>
\end{tcolorbox}

\vspace{1em}

\begin{tcolorbox}[
  enhanced,
  breakable,
  colback=ysshallowgrey,
  colframe=ysdarkgrey,
  title={Point-wise Score Prompt},
  fonttitle=\bfseries
]
<system\_role>

You are a strict, meticulous, and objective research article evaluation expert.

You excel at using specific assessment criteria to thoroughly evaluate research articles, providing precise scores and clear justifications.

</system\_role>

<user\_prompt>

\textbf{Task Background}

There is a deep research task, and you need to evaluate a research article written for this task.

We will assess the article across four dimensions: Comprehensiveness, Insight, Instruction Following, and Readability.

The content is as follows:

<task>

"\{task\_prompt\}"

</task>

\textbf{Article to Evaluate}

<target\_article>

"\{article\}"

</target\_article>

\textbf{Evaluation Criteria}

Now, you need to evaluate this article based on the following \textbf{evaluation criteria list}, providing analysis and scoring each on a scale of 0-10.

Each criterion includes an explanation, please understand carefully.

<criteria\_list>

\{criteria\_list\}

</criteria\_list>

<Instruction>

\textbf{Your Task}

Please strictly evaluate `<target\_article>` based on \textbf{each criterion} in the `<criteria\_list>`.

You need to:
\begin{enumerate}
    \item \textbf{Analyze Each Criterion}: Consider how the article fulfills the requirements of each criterion.
    \item \textbf{Analysis and Evaluation}: Analyze the article's performance on each criterion, referencing the content and criterion explanation, noting strengths and weaknesses.
    \item \textbf{Score}: Based on your analysis, score the article on each criterion (0-10 points).
\end{enumerate}

\textbf{Scoring Rules}

For each criterion, score the article on a scale of 0-10 (continuous values).

The score should reflect the quality of performance on that criterion:
\begin{itemize}
    \item 0-2 points: Very poor performance. Almost completely fails to meet the criterion requirements.
    \item 2-4 points: Poor performance. Minimally meets the criterion requirements with significant deficiencies.
    \item 4-6 points: Average performance. Basically meets the criterion requirements, neither good nor bad.
    \item 6-8 points: Good performance. Largely meets the criterion requirements with notable strengths.
    \item 8-10 points: Excellent/outstanding performance. Fully meets or exceeds the criterion requirements.
\end{itemize}

\textbf{Output Format Requirements}

Please \textbf{strictly} follow the `<output\_format>` below for each criterion evaluation.

\textbf{Do not include any other unrelated content, introduction, or summary}.

Start with "Standard 1" and proceed sequentially through all criteria:

</Instruction>

<output\_format>

\{\{

    "comprehensiveness": [
    
        \{\{
        
            "criterion": [Text content of the first comprehensiveness evaluation criterion],
            
            "analysis": [Analysis],
            
            "target\_score": [Continuous score 0-10]
            
        \}\},
        
        \{\{
        
            "criterion": [Text content of the second comprehensiveness evaluation criterion],
            
            "analysis": [Analysis],
            
            "target\_score": [Continuous score 0-10]
            
        \}\},
        
        ...
        
    ],
    
    "insight": [
    
        \{\{
        
            "criterion": [Text content of the first insight evaluation criterion],
            
            "analysis": [Analysis],
            
            "target\_score": [Continuous score 0-10]
            
        \}\},
        
        ...
        
    ],
    
    ...
    
\}\}

</output\_format>

Now, please evaluate the article based on the research task and criteria, providing detailed analysis and scores according to the requirements above.

Ensure your output follows the specified `<output\_format>` and that the JSON format is parsable, with all characters that might cause JSON parsing errors properly escaped.

</user\_prompt>
\end{tcolorbox}

\newpage

\begin{tcolorbox}[
  enhanced,
  breakable,
  colback=ysshallowgrey,
  colframe=ysdarkgrey,
  title={Vanilla Prompt},
  fonttitle=\bfseries
]
<system\_role>

You are a strict, meticulous, and objective research article evaluation expert.

You excel at using specific assessment criteria to thoroughly evaluate research articles, providing precise scores and clear justifications.

</system\_role>

<user\_prompt>

\textbf{Task Background}

There is a deep research task, and you need to evaluate a research article written for this task.

<task>

"\{task\_prompt\}"

</task>

\textbf{Article to Evaluate}

<target\_article>

"\{article\}"

</target\_article>

<Instruction>

\textbf{Your Task}

Please evaluate the overall quality of the above `<target\_article>` as a response to `<task>`.

Please provide an overall score between 0 and 10.

Also, provide a brief justification for your score.

\textbf{Output Format Requirements}

Please \textbf{strictly} follow the `<output\_format>` below for your evaluation result.

\textbf{Do not include any other unrelated content, introduction, or summary}.

</Instruction>

<output\_format>

\{\{
    "overall\_score": [Continuous score 0-10],
    "justification": "[Scoring justification]"
\}\}

</output\_format>

Now, please evaluate the article based on the task and provide your score and justification according to the specified format.

Ensure your output is valid JSON format and escape any special characters as needed.

</user\_prompt>
\end{tcolorbox}

\vspace{1em}

\end{document}